\documentclass[letterpaper,journal]{IEEEtran}

\usepackage{amsmath,amsfonts}
\usepackage{amssymb}
\usepackage{algorithm}
\usepackage{booktabs}
\usepackage{multirow}
\makeatletter
\renewcommand{\fnum@algorithm}{Algorithm~\thealgorithm:}
\makeatother
\usepackage{microtype}
\usepackage{algpseudocode}
\usepackage{array}
\usepackage[caption=false,font=normalsize,labelfont=sf,textfont=sf]{subfig}
\usepackage{textcomp}
\usepackage{stfloats}
\usepackage{url}
\usepackage{graphicx}
\usepackage{cite}
\usepackage{tabularx}
\usepackage{wasysym}

\usepackage{tabularx}
\usepackage{array}
\usepackage{cellspace}
\usepackage{tikz}
\usepackage{booktabs}
\usepackage{multirow}
\addparagraphcolumntypes{X}

\newcommand{\nomark}{%
    \tikz[baseline=-0.25ex]{
        \draw[line width=0.4pt] (0,0) circle (0.72ex);
    }%
}

\newcommand{\yesmark}{%
    \tikz[baseline=-0.25ex]{
        \fill (0,0) circle (0.72ex);
    }%
}

\newcommand{\partmark}{%
    \tikz[baseline=-0.25ex]{
        \begin{scope}
            \clip (0,0) circle (0.72ex);
            \fill (-0.72ex,-0.72ex)
                  rectangle (0,0.72ex);
        \end{scope}
        \draw[line width=0.4pt] (0,0) circle (0.72ex);
    }%
}
\allowdisplaybreaks
\begin{document}

\title{DyRadio: Spatiotemporal Diffusion for Dynamic Radio Map Reconstruction
}

\author{Zhijie Fan, Zhifeng Zhao and Rongpeng Li
\thanks{Z. Fan and R. Li are with the College of Information Science and Electronic Engineering, Zhejiang University, Hangzhou 310027, China (e-mail: \{fanzhijie, lirongpeng\}@zju.edu.cn).}

\thanks{Z. Zhao is with Zhejiang Lab, Hangzhou 310012, China, and also with the College of Information Science and Electronic Engineering, Zhejiang University, Hangzhou 310027, China (e-mail: zhaozf@zhejianglab.org).}}

\maketitle


\begin{abstract}
Accurate radio map (RM) reconstruction
is essential for environment-aware wireless networks.
Sampling-based methods, which infer radio conditions
from sparse measurements, are widely adopted because they do not require explicit knowledge of transmitter configurations.
However, in dynamic environments, time-varying propagation
conditions cause RMs to evolve, complicating
reconstruction from temporally distributed sparse observations.
Although cross-frame observations provide complementary
spatial information, their direct aggregation can introduce
temporal state mismatch. Furthermore, unknown measurement
noise adds to the reconstruction difficulties.
To address these challenges, we propose DyRadio,
a framework that integrates spatiotemporal RM reconstruction
with unknown noise estimation.
Specifically, a propagation-prior-guided,
noise-conditioned spatiotemporal diffusion model (DM)
jointly reconstructs consecutive RMs by capturing cross-frame dependencies, while adapting to measurement noise levels.
To estimate the noise level, an iterative
predictive likelihood method alternates between DM-based blind-area prediction and noise-level updates.
The estimated noise level then conditions reconstruction
of the complete RM sequence.
Experimental results demonstrate superior reconstruction
accuracy and robustness against measurement noise,
with peak signal-to-noise ratio (PSNR) improvements of $1.18$--$5.75$ dB
over the best-performing baseline across all evaluated settings.

\end{abstract}


\section{Introduction}
\label{sec:introduction}

Next-generation (xG) wireless networks are expected to integrate communications with intelligence and sensing capabilities ~\cite{dang2020what,letaief2019roadmap}, so as to better
understand the surrounding physical environment
and its impact on radio propagation. Recent advances in digital twins and environment-aware
wireless intelligence further highlight the importance
of accurately characterizing radio conditions
across physical spaces~\cite{chen2026wireless,kumbhar2026physicalai,wang2025digital}. A radio map (RM), which characterizes the spatial distribution
of radio quantities (e.g., received signal strength)
across a target area, could benefit spectrum management~\cite{bi2019engineering}, interference-aware predictive communication~\cite{li2024radiomap}, and proactive handover based on anticipated signal quality~\cite{rakkesh2026handover}. However, most existing literature focuses on single-frame RM reconstruction~\cite{wang2025radiodiff,wang2025radiodiffk2}, while, along with the growing number of vehicular, robotic, and agentic applications~\cite{roger2024v2x,li2024radiomap,wen2026bridging}, there emerges a strong incentive toward developing dynamic RM reconstruction methods. 
\subsection{Related Works}
Conventional RM reconstruction generally relies on the spatial interpolation of field measurements or physics-based simulations~\cite{romero2022radiomap}. However, the spatial interpolation techniques, such as radial basis functions, splines~\cite{mateos2009spline}, and Kriging~\cite{dallanese2011kriging}, suffer from deteriorated reconstruction accuracy when measurements are sparse or propagation conditions are complex. Meanwhile, physics-based simulation methods, including ray tracing~\cite{hoydis2023sionna}, often induce substantial computational overhead, particularly in complex and time-varying environments.

These limitations have motivated the adoption of learning-based RM reconstruction methods. Among them, sampling-free approaches rely on explicit transmitter
information, including known transmitter (TX) locations and
configurations, together with environmental information,
to construct RMs. Early learning-based methods, such as RadioUNet~\cite{levie2021radiounet} and RadioNet~\cite{tian2021transformer}, employ convolutional networks and Transformer architectures to learn the deterministic mapping from transmitter information and environmental layouts to spatial radio fields. More recently, generative approaches have been introduced to capture complex propagation characteristics beyond deterministic predictions. By employing a diffusion model (DM)~\cite{ho2020denoising},
RadioDiff~\cite{wang2025radiodiff} generates high-fidelity
RMs in a sampling-free manner, while
RadioDiff-$k^2$~\cite{wang2025radiodiffk2} further
incorporates electromagnetic propagation characteristics
to improve multipath modeling. Nevertheless, the required information, such as transmitter locations
and transmission configurations, may be unavailable
in practical deployments due to privacy restrictions or
non-cooperative transmitters.

On the other hand, sampling-based approaches leverage
sparse radio measurements and can reduce reliance
on explicit transmitter information. Initially, learning-based methods utilize deep completion
autoencoders~\cite{teganya2021deep} or adversarial
learning~\cite{zhang2023rmegan} to estimate RMs
from sparse observations. Recent advances in DM have further improved the capability of recovering propagation patterns under limited measurements. In particular, WiFi-Diffusion~\cite{liu2025wifidiffusion} employs a DM to
generate multiple candidate RMs and selects the
final reconstruction through a propagation-guided
evaluation mechanism, achieving accurate RM
reconstruction at ultra-low sampling rates. Physics-informed reconstruction methods, such as PhyRMDM~\cite{jia2025physics}, further incorporate electromagnetic constraints to improve the consistency of reconstructed RMs with propagation characteristics. RadioDiff-Inverse~\cite{wang2026radiodiffinverse} formulates RM reconstruction from noisy sparse measurements as a Bayesian inverse problem, leveraging a DM-based prior to recover RM under incomplete environmental information. Although these approaches have substantially advanced sparse RM reconstruction, they primarily consider individual, single-frame RM reconstruction and do not explicitly exploit the dependencies among temporally distributed observations in continuously evolving environments.

\begin{figure*}[!t]
    \centering
    \includegraphics[width=0.95\textwidth]{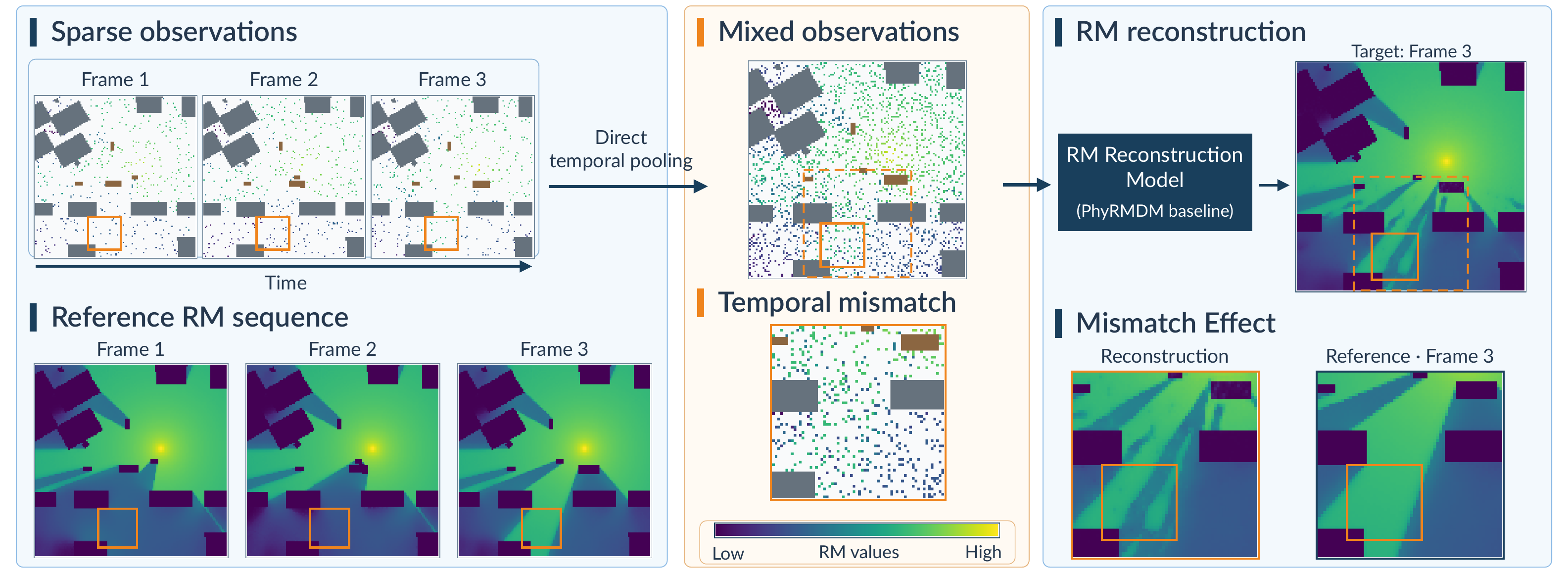}
    \caption{Illustration of temporal state mismatch in dynamic RM reconstruction.}
    \label{fig:motivation}
\end{figure*}

In dynamic wireless environments, RMs vary over time
as propagation conditions evolve. Consequently, previously constructed RMs may become
outdated, motivating research on dynamic RM reconstruction methods.
RadioMapMotion~\cite{cheng2026radiomapmotion}
employs ConvLSTM to predict future RM sequences
from historical complete RMs.
However, acquiring dense measurements at every
time incurs substantial overhead,
leaving only sparse observations.
RadioProphet~\cite{shao2026radioprophet}
further investigates future RM prediction
from historical sparse observations through
collaboration between edge-side single-frame reconstruction and cloud-side prediction models. Updating changing RMs with sparse observations is essential
to reflect dynamic radio conditions.
However, existing prediction methods mainly fuse
historical RM features for prediction, leaving the challenge of
exploiting complementary observations across frames while
preserving their distinct
propagation states.

Here, we showcase the joint reconstruction
of a sequence of dynamic RMs from temporally
distributed sparse observations in urban traffic scenarios. Specifically, moving vehicles alter
signal blockage, reflection, and multipath
propagation, causing RM values to change over time. 
Fig.~\ref{fig:motivation} illustrates RM reconstruction results through
direct temporal pooling in PhyRMDM~\cite{jia2025physics}, where only building layout
and time-varying vehicle configurations are utilized
but no explicit transmitter information is taken into account.
It can be observed that directly combining observations from different frames
ignores changes in propagation conditions, and the
resulting temporal state mismatch leads to distorted
propagation patterns near moving vehicles, which is termed \emph{temporal state mismatch}.
Beyond temporal state mismatch, observation noise poses an additional challenge to spatiotemporal RM reconstruction. In dynamic environments, both observation noise and changes in propagation conditions cause variations in observed RMs. Distinguishing these effects becomes difficult when only sparse observations are available at each instant. Existing probabilistic methods, such as RadioDiff-Inverse~\cite{wang2026radiodiffinverse}, account for measurement noise using a likelihood model parameterized by an assumed known noise variance. However, the actual noise level is often unknown in practice. Reconstruction errors also contribute to the discrepancies between observations and predicted RM, complicating noise-level estimation. Therefore, estimating the unknown noise level while accounting for prediction uncertainty is important for subsequent noise-conditioned spatiotemporal RM reconstruction.
\subsection{Contributions}
To address these challenges, we propose DyRadio,
a framework for dynamic RM reconstruction
with unknown measurement noise estimation. Guided by propagation prior features from a U-Net~\cite{ronneberger2015unet},
we develop a spatiotemporal DM based on
the Diffusion Transformer (DiT) architecture~\cite{peebles2023dit}.
Through 3D patch embedding and spatiotemporal
self-attention, the DiT captures spatial structures
and cross-frame dependencies for joint RM reconstruction while mitigating temporal mismatch.
Inspired by previous studies~\cite{ho2022cascaded,heurteldepeiges2024listening},
we introduce a noise level-modulated cross-attention
module to integrate sparse RM observations
into the generation process. By injecting the measurement noise level,
the proposed DM can adapt to varying
noise conditions. Furthermore, since the measurement noise level is typically unknown in practice, we propose an iterative method based on blind-area predictive likelihood, which alternates between reconstructing partial RMs
using the proposed DM and estimating \& updating the noise level. This enables noise level estimation from sparse
observations without access to complete RMs,
while the estimated noise level is then used to condition the proposed DM to reconstruct the full RM sequence. 
\begin{table*}[!t]
\centering
\caption{Comparison with Representative Studies on Radio Map Modeling.}
\label{tab:literature_comparison}
\footnotesize

\setlength{\tabcolsep}{2.2pt}
\setlength{\cellspacetoplimit}{3pt}
\setlength{\cellspacebottomlimit}{3pt}
\setlength{\extrarowheight}{0.5pt}
\renewcommand{\arraystretch}{1.12}
\setlength{\arrayrulewidth}{0.35pt}
\renewcommand{\tabularxcolumn}[1]{m{#1}}

\begin{tabularx}{\textwidth}{@{}
>{\raggedright\arraybackslash}S{m{0.150\textwidth}}|
>{\centering\arraybackslash}S{m{0.080\textwidth}}
>{\centering\arraybackslash}S{m{0.080\textwidth}}
>{\centering\arraybackslash}S{m{0.078\textwidth}}
>{\centering\arraybackslash}S{m{0.085\textwidth}}
>{\centering\arraybackslash}S{m{0.096\textwidth}}
>{\centering\arraybackslash}S{m{0.096\textwidth}}|
>{\raggedright\arraybackslash}S{X}
@{}}

\noalign{\hrule height 0.8pt}

Representative Studies
& \shortstack{Learning-\\Based}
& \shortstack{Diffusion-\\Based}
& \shortstack{TX-Free\\Inference}
& \shortstack{Temporal\\Modeling}
& \shortstack{Dynamic RM\\Reconstruction}
& \shortstack{Noise-Adaptive\\Reconstruction}
& Brief Description \\

\hline

\cite{mateos2009spline,dallanese2011kriging,hoydis2023sionna}
& \nomark & \nomark & \partmark & \nomark & \nomark & \nomark
& Limited accuracy under sparse sampling; high simulation cost \\

\hline

\cite{levie2021radiounet,tian2021transformer,teganya2021deep,zhang2023rmegan}
& \yesmark & \nomark & \partmark & \nomark & \nomark & \nomark
& Limited reconstruction fidelity without diffusion-based generation \\

\hline

\cite{wang2025radiodiff,wang2025radiodiffk2}
& \yesmark & \yesmark & \nomark & \nomark & \nomark & \nomark
& Requires explicit transmitter information \\

\hline

\cite{liu2025wifidiffusion,jia2025physics,wang2026radiodiffinverse}
& \yesmark & \yesmark & \partmark & \nomark & \nomark & \partmark
& Unable to model temporal dynamics \\

\hline

\cite{cheng2026radiomapmotion,shao2026radioprophet}
& \yesmark & \nomark & \yesmark & \yesmark & \nomark & \nomark
& Focuses on history-driven prediction rather than sparse-observation reconstruction \\

\hline

\textbf{Ours}
& \yesmark & \yesmark & \yesmark & \yesmark & \yesmark & \yesmark
& Jointly reconstructs dynamic RM sequences from sparse observations while accounting for temporal mismatch and estimating the unknown measurement-noise level \\

\noalign{\hrule height 0.8pt}

\end{tabularx}

\vspace{5pt}
\noindent
\parbox{\textwidth}{%
    \footnotesize
    \textit{Notations:}~
    \nomark~not included;\quad
    \yesmark~fully supported;\quad
    \partmark~partially supported.
}

\end{table*}
While summarizing the key differences in Table \ref{tab:literature_comparison}, the main contributions of this work are summarized as follows:
\begin{itemize}
    \item 
    We formulate dynamic RM reconstruction from sparse
    observations as a spatiotemporal reconstruction problem.
    To exploit complementary cross-frame information while
    addressing temporal state mismatch, we propose a
    propagation-guided and noise-conditioned spatiotemporal DM.
    Furthermore, a noise-modulated cross-attention module
    is adopted to enhance reconstruction robustness under
    varying noise conditions.

    \item 
    We develop an iterative predictive likelihood method
to estimate the unknown measurement noise level from sparse observations. By exploiting the learned RM prior of the proposed DM,
the method alternately refines blind-area RM predictions
and updates the noise estimate without requiring
complete RMs or an additional estimation network. The resulting estimate provides a measure
of observation uncertainty, enabling adaptive RM
reconstruction under varying noise conditions.
\item 
To validate the proposed framework, we construct
a dynamic urban traffic RM dataset using
CARLA~\cite{dosovitskiy2017carla} and Sionna
RT~\cite{hoydis2023sionna}, enabling realistic
vehicle motion and time-varying radio propagation.
Extensive experiments across different sampling rates
and noise levels demonstrate that DyRadio outperforms
existing RM reconstruction methods\cite{wang2025radiodiff, jia2025physics, levie2021radiounet, cheng2026radiomapmotion}, achieving PSNR
improvements of $1.18$--$5.75$ dB over the strongest
baseline. 
\end{itemize}

\subsection{Paper Structure}
The remainder of this paper is organized as follows.
Section~\ref{sec:preliminaries} first introduces the diffusion
model preliminaries and formulates the dynamic RM reconstruction
problem.
Section~\ref{sec:method} then presents the proposed
spatiotemporal DM and iterative noise estimation method.
Section~\ref{sec:experiments} provides extensive experimental
results to demonstrate the reconstruction advantages of DyRadio
and evaluate the accuracy of noise estimation.
Finally, Section~\ref{sec:conclusion} concludes the paper.
\section{Preliminary and Problem Formulation}
\label{sec:preliminaries}
This section first introduces the diffusion modeling preliminaries and then formulates the dynamic RM reconstruction problem under unknown measurement noise.

\subsection{Diffusion Model Preliminaries}
\label{diffusion_model_preliminaries}
Diffusion models learn to generate data through a sequence of
forward noising and reverse denoising~\cite{ho2020denoising}. The forward process
gradually adds Gaussian noise to a clean sample, while a learned
reverse process recovers data starting from Gaussian noise.
For conditional generation, the reverse process additionally
uses available information $\mathbf{c}$ to generate samples
from $p_\theta(\mathbf{z}_0\mid\mathbf{c})$, where
$\mathbf{z}_0$ denotes the clean data
sample~\cite{saharia2022image}. The condition is provided to
the denoising network during both training and sampling.

Starting from  $\mathbf z_0$, the forward process
gradually adds Gaussian noise to obtain noisy sample $\mathbf z_\tau$
through a Markov chain:
\begin{equation}
q(\mathbf z_\tau\mid\mathbf z_{\tau-1})
=\mathcal{N}(\sqrt{\alpha_\tau}\mathbf z_{\tau-1},
\beta_\tau\mathbf I),
\end{equation}
where $\tau\in\{1,\ldots,\mathcal T\}$,
$\beta_\tau$ is the prescribed noise variance,
and $\alpha_\tau=1-\beta_\tau$. Defining
$\bar{\alpha}_\tau=\prod_{j=1}^{\tau}\alpha_j$ and $\bar{\alpha}_0=1$, a noisy sample at
an arbitrary step can be obtained directly from
$\mathbf{z}_0$ as
\begin{equation}
    \mathbf{z}_\tau
    =
    \sqrt{\bar{\alpha}_\tau}\mathbf{z}_0
    +
    \sqrt{1-\bar{\alpha}_\tau}\boldsymbol{\epsilon},
    \qquad
    \boldsymbol{\epsilon}\sim\mathcal{N}(\mathbf{0},\mathbf{I}).
    \label{eq:forward_diffusion}
\end{equation}
 With
$\bar{\alpha}_\tau$ sufficiently close to zero, the terminal
distribution is approximately standard Gaussian.

Generation requires the reverse transition
$q(\mathbf{z}_{\tau-1}\mid\mathbf{z}_\tau,\mathbf{c})$, which depends
on the unknown conditional data distribution. During training,
however, the clean sample $\mathbf{z}_0$ is available.
Conditioning additionally on this sample yields a tractable
Gaussian posterior. Since the forward process is independent
of $\mathbf{c}$ given $\mathbf{z}_0$,
for $\tau\geq 2$,
\begin{equation}
\begin{aligned}
    q(\mathbf{z}_{\tau-1}\mid\mathbf{z}_\tau,\mathbf{z}_0)
    &=
    \frac{
        q(\mathbf{z}_\tau\mid\mathbf{z}_{\tau-1})
        q(\mathbf{z}_{\tau-1}\mid\mathbf{z}_0)
    }{
        q(\mathbf{z}_\tau\mid\mathbf{z}_0)
    } \\
    &=
    \mathcal{N}\!\left(
        \mathbf{z}_{\tau-1};
        \tilde{\boldsymbol{\mu}}_\tau,
        \tilde{\beta}_\tau\mathbf{I}
    \right),
    \label{eq:diffusion_posterior}
\end{aligned}
\end{equation}
where $\tilde{\boldsymbol{\mu}}_\tau$ is the posterior mean and
$\tilde{\beta}_\tau
=\beta_\tau(1-\bar{\alpha}_{\tau-1})/(1-\bar{\alpha}_\tau)$.

Following the denoising diffusion probabilistic model (DDPM) formulation, the learned reverse transition is modeled as
a Gaussian with mean
$\boldsymbol{\mu}_\theta(\mathbf{z}_\tau,\tau,\mathbf{c})$
and a fixed covariance schedule $\boldsymbol{\Sigma}_\tau$.
Using Eq.~(\ref{eq:forward_diffusion}) and replacing the added noise
$\boldsymbol{\epsilon}$ with the neural network prediction
$\boldsymbol{\epsilon}_\theta(\mathbf{z}_\tau,\tau,\mathbf{c})$,
the reverse-process mean $\boldsymbol{\mu}_\theta$ can be expressed as
\begin{equation}
    \boldsymbol{\mu}_\theta(\mathbf{z}_\tau,\tau,\mathbf{c})
    =
    \frac{1}{\sqrt{\alpha_\tau}}
    \left(
        \mathbf{z}_\tau
        -
        \frac{\beta_\tau}{\sqrt{1-\bar{\alpha}_\tau}}
        \boldsymbol{\epsilon}_\theta(\mathbf{z}_\tau,\tau,\mathbf{c})
    \right).
    \label{eq:reverse_mean}
\end{equation}
Under this parameterization, the intermediate terms of the
variational training objective reduce to weighted
noise-prediction errors. The commonly used simplified
objective omits the step-dependent weights:
\begin{equation}
    \mathcal{L}_{\mathrm{diff}}
    =
    \mathbb{E}_{(\mathbf{z}_0,\mathbf{c}),\tau,\boldsymbol{\epsilon}}
    \left[
        \left\|
            \boldsymbol{\epsilon}
            -
            \boldsymbol{\epsilon}_\theta(
                \mathbf{z}_\tau,\tau,\mathbf{c}
            )
        \right\|_2^2
    \right],
    \label{eq:diffusion_loss}
\end{equation}
where $(\mathbf{z}_0,\mathbf{c})$ is drawn from the training
data and $\tau$ is sampled uniformly from $\{1,\ldots,\mathcal T\}$.
The network therefore learns to estimate the added noise at
different diffusion steps using both the noisy sample and
its conditioning information.

Standard DDPM sampling traverses all diffusion steps,
requiring substantial computational overhead. Denoising diffusion
implicit model (DDIM)~\cite{song2021denoising} enables accelerated sampling by denoising over
a selected subsequence of diffusion steps. 
In deterministic DDIM sampling, the predicted noise first
provides an estimate of the clean sample from $\mathbf{z}_{\tau_1}$ by rearranging
Eq. \eqref{eq:forward_diffusion}
\begin{equation}
\begin{aligned}
    \hat{\mathbf{z}}_0^{(\tau_1)}
    &=
    \frac{
        \mathbf{z}_{\tau_1}
        -
        \sqrt{1-\bar{\alpha}_{\tau_1}}
        \boldsymbol{\epsilon}_\theta(
            \mathbf{z}_{\tau_1},\tau_1,\mathbf{c}
        )
    }{
        \sqrt{\bar{\alpha}_{\tau_1}}
    }.
\end{aligned}
\label{eq:ddim_sampling}
\end{equation}
The clean-sample estimate and
the predicted noise are then combined according to the
noise level of the next selected step, where $\tau_2<\tau_1$ denotes the next selected step:
\begin{equation}
\begin{aligned}
    \mathbf{z}_{\tau_2}
    =
    \sqrt{\bar{\alpha}_{\tau_2}}
    \hat{\mathbf{z}}_0^{(\tau_1)}
    +
    \sqrt{1-\bar{\alpha}_{\tau_2}}
    \boldsymbol{\epsilon}_\theta(
        \mathbf{z}_{\tau_1},\tau_1,\mathbf{c}
    ).
\end{aligned}
\label{eq:ddim_denoise}
\end{equation}
Unlike stochastic DDPM sampling, deterministic DDIM
introduces no additional random noise during reverse updates.
Thus, the sampling trajectory is fully determined by the
initial Gaussian sample and the condition $\mathbf{c}$.

\subsection{Problem Formulation}
Consider a dynamic wireless propagation environment over a spatial region discretized into $H \times W$ grid points. Let $\mathbf{x}_t \in \mathbb{R}^{HW}$ denote the vectorized RM at time $t$, representing the spatial distribution of radio propagation quantities. The complete RM sequence over $T$ consecutive time instants is denoted by $\mathbf{X} = \{\mathbf{x}_1, \mathbf{x}_2, \ldots, \mathbf{x}_T\}$. As the propagation environment evolves, consecutive radio maps exhibit distinct yet temporally correlated propagation states.

In wireless systems, radio measurements are available only at a limited number of spatial locations at each time instant. Let $\mathbf{A}_t \in \{0,1\}^{m_t \times HW}$ denote the sampling matrix that selects $m_t$ observed locations from $\mathbf{x}_t$ at a sampling rate of $p$. The acquired measurements are modeled as
\begin{equation}
    \mathbf{y}_t = \mathbf{A}_t \mathbf{x}_t + \mathbf{n}_t,
    \quad
    \mathbf{n}_t \sim \mathcal{N}(\mathbf{0}, \sigma_{\mathrm{obs}}^2 \mathbf{I}_{m_t}),
    \label{eq:observation_model}
\end{equation}
where $\mathbf{y}_t \in \mathbb{R}^{m_t}$ contains the observed values, and $\sigma_{\mathrm{obs}}$ denotes the standard deviation of measurement noise.

Our objective is to reconstruct the complete RM sequence
$\mathbf{X}$ from temporally distributed sparse measurements
and environmental information. Given the measurement noise
level $\sigma_{\mathrm{obs}}$, the reconstruction objective is formulated as
\begin{equation}
\max_{\mathbf{X}}\;
p(\mathbf{X}\mid\mathcal{Y},\mathcal{A},\mathbf{C},\sigma_{\mathrm{obs}}),
\label{eq:conditional_reconstruction}
\end{equation}
where $\mathcal{Y}=\{\mathbf{y}_t\}_{t=1}^{T}$
and $\mathcal{A}=\{\mathbf{A}_t\}_{t=1}^{T}$
denote the measurements and sampling operators,
respectively. The environmental information
$\mathbf{C}=\{\mathbf{B},\mathbf{V}_1,\ldots,\mathbf{V}_T\}$
consists of the static building layout $\mathbf{B}$
and vehicle configurations $\mathbf{V}_t$.
Joint reconstruction exploits complementary measurements
across time while accounting for evolving propagation states.

However, the measurement noise level $\sigma_{\mathrm{obs}}$ is generally unknown in practice, despite its importance for reconstruction in inverse problems~\cite{chung2023diffusion,heurteldepeiges2024listening}. We formulate noise-level estimation as maximizing its posterior probability:
\begin{equation}
\max_{\sigma_{\mathrm{obs}}\geq0}\;
p(\sigma_{\mathrm{obs}}\mid\mathcal{Y},\mathcal{A},\mathbf{C}).
\end{equation}
Since both the RM reconstruction of $\mathbf{X}$ and estimation of noise level $\sigma_{\mathrm{obs}}$ are unknown, their inference is coupled through the joint posterior:
\begin{equation}
\begin{aligned}
&p(\mathbf{X},\sigma_{\mathrm{obs}}\mid\mathcal{Y},\mathcal{A},\mathbf{C})\\
&=p(\mathbf{X}\mid\mathcal{Y},\mathcal{A},\mathbf{C},\sigma_{\mathrm{obs}})
p(\sigma_{\mathrm{obs}}\mid\mathcal{Y},\mathcal{A},\mathbf{C}).
\end{aligned}
\end{equation}
We address this coupling by developing a noise-conditioned spatiotemporal DM whose predictive distribution is also used to estimate the unknown measurement-noise level. The resulting estimate subsequently conditions the reconstruction of the entire $T$-length RM sequence in Eq. (\ref{eq:conditional_reconstruction}).

\begin{table}[!t]
\centering
\caption{Summary of Key Notations.}
\label{tab:notations}
\footnotesize
\renewcommand{\arraystretch}{1.1}
\setlength{\tabcolsep}{2.5pt}
\begin{tabular}{@{}p{0.39\columnwidth}p{0.57\columnwidth}@{}}
\hline
\textbf{Symbol} & \textbf{Definition} \\
\hline

$\mathbf x_t, \mathbf X$ &
RM at time $t$ and the complete RM sequence. \\

$T, L$ &
Lengths of the complete sequence and each reconstructed subsequence. \\

$\mathbf X_{1:L},
\mathbf X_{1:L}^{(\tau)}$ &
Clean RM subsequence and its noisy diffusion state at step $\tau$. \\

$\mathbf A_t, \mathcal A$ &
Sampling matrix at time $t$ and the collection of sampling matrices. \\

$\mathbf y_t, \mathcal Y$ &
Sparse measurements at time $t$ and their collection across the sequence. \\

$\mathbf B, \mathbf V_t, \mathbf C$ &
Building layout, vehicle configuration at time $t$, and scene information. \\

$p$ & RM sampling rate at each time. \\

$\sigma_{\mathrm{obs}},
\hat{\sigma}_{\mathrm{obs}}$ &
Measurement-noise standard deviation and its final estimate. \\

$\tau, \boldsymbol\epsilon$ &
Diffusion step and Gaussian diffusion noise. \\

$\alpha_\tau, \beta_\tau$ &
Diffusion coefficient and noise variance at step $\tau$. \\

$U_\phi, D_\theta$ &
Stage I U-Net and Stage II noise-conditioned DiT. \\

$\widetilde{\mathbf X}_{1:L},
\mathcal F$ &
Auxiliary RM prediction and propagation features from Stage I. \\

$\mathbf e_\tau, \mathbf e_\sigma$ &
Embeddings of the diffusion step and measurement-noise level. \\

$\Omega_{\mathrm{sam}},
\Omega_{\mathrm{sam}}^{(k)}$ &
Observed locations and those assigned to the $k$-th blind area. \\

$\mathcal Y_{-\Omega_{\mathrm{sam}}^{(k)}}$ &
Observations excluding the $k$-th blind area. \\

$K, S$ &
Numbers of blind areas and Monte Carlo samples per blind area. \\

$r, \omega$ &
Noise-estimation iteration and spatiotemporal location index. \\

$\hat{\mu}_\omega^{(r,k)},
\hat{v}_\omega^{(r,k)}$ &
Monte Carlo estimates of the predictive mean and variance. \\

$\sigma_{\mathrm{obs}}^{(r)},
\widetilde{\sigma}_{\mathrm{obs}}^{(r+1)}$ &
Current noise estimate and the candidate estimate obtained by likelihood optimization. \\

$\eta$ &
Relaxation parameter for iterative noise estimation. \\

$\delta, R_{\max}$ &
Convergence tolerance and maximum number of noise-estimation iterations. \\

\hline
\end{tabular}
\end{table}

\begin{figure}[!t]
    \centering
    \includegraphics[width=\columnwidth]{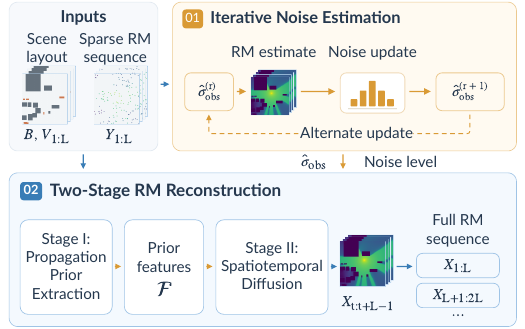}
    \caption{Overview of the proposed framework for noise estimation and spatiotemporal RM reconstruction.}
    \label{fig:framework}
\end{figure}
\begin{figure*}[t]
    \centering
    \includegraphics[
        width=\textwidth,
        height=10cm,
        keepaspectratio
    ]{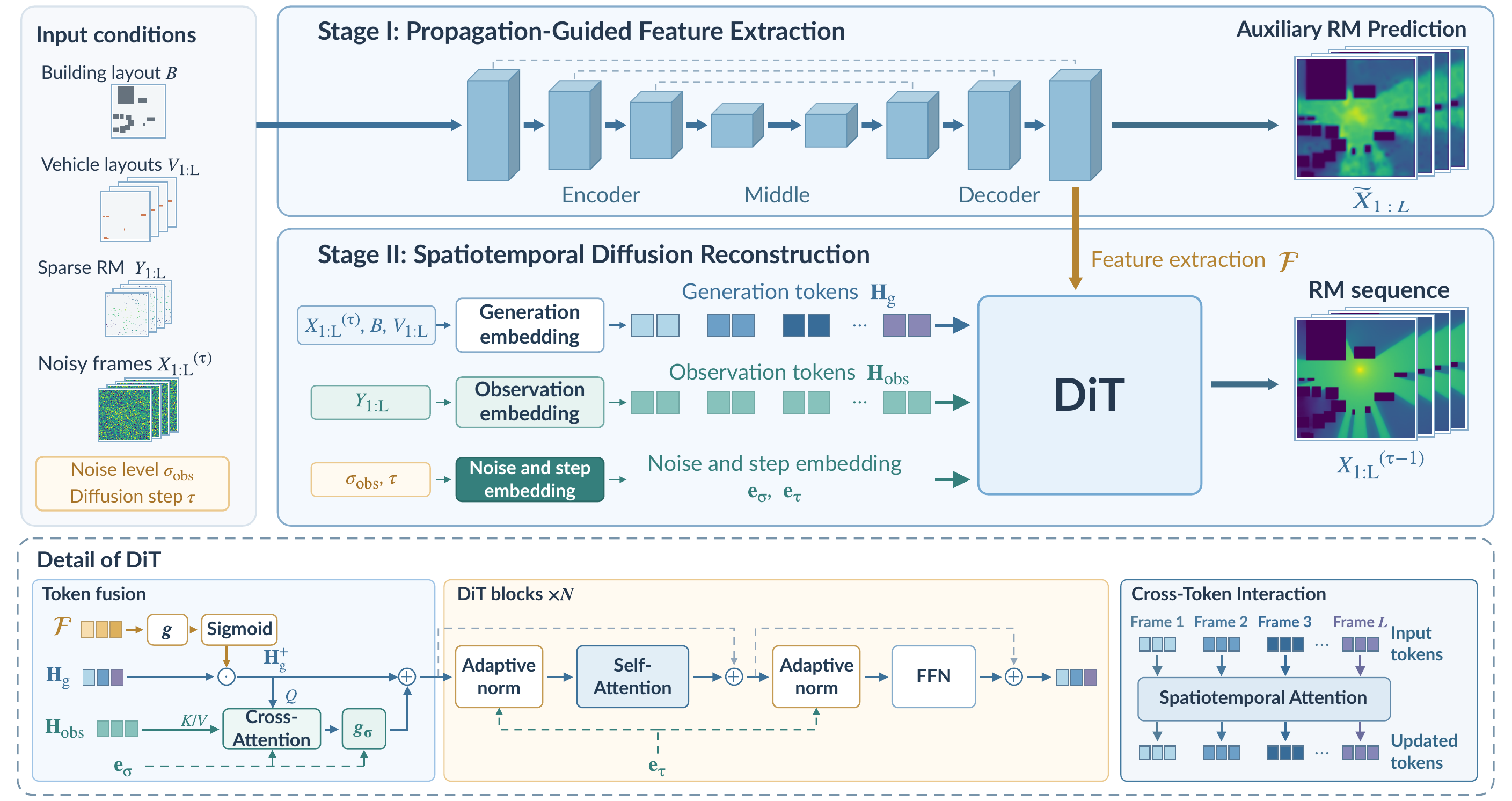}
    \caption{Architecture of the noise-conditioned spatiotemporal DM for dynamic RM reconstruction.}
    \label{fig:model_architecture}
\end{figure*}
\section{The Proposed Method}
\label{sec:method}
\subsection{Framework Overview}
DyRadio reconstructs the complete $T$-frame RM sequence
from scene information and sparse noisy observations.
To limit the computational cost of jointly processing all $T$ frames,
it gradually reconstructs the sequence by one $L$-frame subsequence
at a time. Fig.~\ref{fig:framework} illustrates the overall framework,
which consists of noise estimation followed by RM reconstruction.

Following prior-guided DM designs~\cite{jia2025physics,liu2025wifidiffusion},
we adopt a two-stage architecture to guide diffusion
reconstruction with propagation priors, by first assuming that measurement noise $\sigma_{\mathrm{obs}}$ is known. 
Stage I uses a U-Net to predict a coarse RM and extract
propagation-related features from the final decoder layers.
Stage II employs a noise-conditioned spatiotemporal DiT
to jointly reconstruct each subsequence using these features,
3D patch embeddings, and spatiotemporal self-attention.
This design exploits complementary cross-frame observations
while accounting for temporal state mismatch.
Furthermore, the measurement-noise level modulates observation
cross-attention to adjust the contribution of sparse observations.

To estimate the measurement noise $\sigma_{\mathrm{obs}}$, we further adopt an iterative blind-area predictive
likelihood method.
The method alternates between blind-area RM prediction
with the pretrained DM and predictive-likelihood updates
of the noise estimate, accounting for predictive uncertainty.
Assuming a shared noise level across the whole $T$-frame sequence,
the final estimate $\hat{\sigma}_{\mathrm{obs}}$
conditions the reconstruction of all following $L$-frame subsequences.

\subsection{Noise-Conditioned Spatiotemporal DM}
Fig.~\ref{fig:model_architecture} illustrates the architecture
of the proposed two-stage DM.
We first describe the propagation-guided feature extraction
in Stage I, and then present the noise-conditioned
spatiotemporal reconstruction in Stage II.
\subsubsection{Stage I: Propagation-Guided Feature Extraction}

The first stage learns propagation-prior features through
auxiliary RM prediction. A U-Net-based model
$U_{\phi}$ takes the scene information and sparse RM
observations as inputs, producing an auxiliary RM prediction
$\widetilde{\mathbf X}_{1:L}$ and propagation-prior features
$\mathcal F$:
\begin{equation}
(\widetilde{\mathbf X}_{1:L},\mathcal F)
=U_{\phi}\big(\mathbf C_{1:L},\mathcal Y_{1:L}\big).
\label{eq:auxiliary_prediction}
\end{equation}
Here, $\mathcal F$ collects feature maps extracted from the second- and third-last decoder layers of $U_{\phi}$, with resolutions of $64\times64$ and $32\times32$. These features
are then passed to Stage II to guide spatiotemporal RM
reconstruction.

The auxiliary RM prediction is supervised by a mean squared
error (MSE) loss over all spatial locations and frames:
\begin{equation}
\mathcal L_{\mathrm{MSE}}
= \frac{1}{LHW}
\sum_{l=1}^{L}
\sum_{i=1}^{HW}
(\widetilde{x}_{l,i}-x_{l,i})^2,
\label{eq:prior_mse}
\end{equation}
where $\widetilde{x}_{l,i}$ and $x_{l,i}$ denote the
predicted and ground-truth RM values at location $i$
in frame $l$, respectively.

Obstacle locations indicate propagation blockage, while
transmitter locations provide supervision for learning
source-related spatial patterns. We incorporate these
propagation priors into auxiliary RM prediction through
location-specific losses, without requiring transmitter
locations as inference inputs:
\begin{align}
\mathcal L_{\mathrm{occ}}
&= \frac{1}{N_{\mathrm{occ}}}
\sum_{l=1}^{L}
\sum_{i\in\Omega_{\mathrm{occ},l}}
(\widetilde{x}_{l,i}-u_{\mathrm{occ}})^2,
\label{eq:prior_obstacle}\\
\mathcal L_{\mathrm{src}}
&= \frac{1}{N_{\mathrm{src}}}
\sum_{l=1}^{L}
\sum_{i\in\Omega_{\mathrm{src},l}}
(\widetilde{x}_{l,i}-u_{\mathrm{src}})^2.
\label{eq:prior_source}
\end{align}
where $\Omega_{\mathrm{occ},l}$ and
$\Omega_{\mathrm{src},l}$ denote the obstacle and transmitter
locations in frame $l$, respectively.
$N_{\mathrm{occ}}$ and $N_{\mathrm{src}}$ are the corresponding
total numbers of locations across the $L$ frames. $u_{\mathrm{occ}}$ and $u_{\mathrm{src}}$ denote the
predefined fixed RM target values for obstacle and
transmitter locations.
The transmitter locations are used only to compute
$\mathcal L_{\mathrm{src}}$ during training.

The obstacle and source losses complement global MSE
supervision by incorporating scene-dependent
propagation priors. The overall auxiliary objective is
\begin{equation}
\begin{aligned}
\mathcal L_{\mathrm{aux}}
=\ &\lambda_{\mathrm{MSE}}\mathcal L_{\mathrm{MSE}}
+ \lambda_{\mathrm{occ}}\mathcal L_{\mathrm{occ}}
+ \lambda_{\mathrm{src}}\mathcal L_{\mathrm{src}},
\end{aligned}
\label{eq:prior_total}
\end{equation}
where $\lambda_{\mathrm{MSE}}$, $\lambda_{\mathrm{occ}}$,
and $\lambda_{\mathrm{src}}$ are the corresponding loss weights.

\subsubsection{Stage II: Spatiotemporal Diffusion Reconstruction}

Following the noise-prediction formulation introduced in
Section \ref{diffusion_model_preliminaries}, Stage II employs a noise-conditioned DiT
$D_{\theta}$ to predict the diffusion noise from a noisy
$L$-frame RM subsequence and its conditioning information:
\begin{equation}
\begin{aligned}
\hat{\boldsymbol{\epsilon}}
=D_{\theta}\big(
\mathbf X_{1:L}^{(\tau)},\tau,\mathbf{C}_{1:L},
\mathcal Y_{1:L},
\sigma_{\mathrm{obs}},\mathcal F
\big).
\end{aligned}
\label{eq:dit_prediction}
\end{equation}
Here, the conditioning information $\mathbf c$, introduced
in Section II-A, consists of the scene information
$\mathbf C_{1:L}$, sparse RM observations $\mathcal{Y}_{1:L}$,
measurement-noise level $\sigma_{\mathrm{obs}}$, and
propagation-prior features $\mathcal F$. During inference, reverse DDIM sampling starts from Gaussian
noise and uses the predicted diffusion noise at each step to
recover the clean RM subsequence.

In the proposed model $D_\theta$, the noisy diffusion state $\mathbf X_{1:L}^{(\tau)}$ and scene
information $\mathbf C_{1:L}$ are encoded into generation
tokens $\mathbf H_{\mathrm g}$. Meanwhile, sparse RM observations $\mathcal Y_{1:L}$
are encoded into observation tokens $\mathbf H_{\mathrm {obs}}$ using 3D patch embeddings.
To guide generation with propagation priors, we fuse the
features $\mathcal F$ extracted by Stage I into the generation
features through multiplicative feature guidance:
\begin{equation}
\mathbf H_{\mathrm g}^{+}
=
\mathbf H_{\mathrm g}\odot
\operatorname{Sigmoid}\left(g(\mathcal F)\right),
\label{eq:feature_guidance}
\end{equation}
where $\mathbf H_{\mathrm g}^{+}$ denotes
the generation features after fusion
and $g(\cdot)$ matches the channel dimension of $\mathcal F$ and $\mathbf H_{\mathrm g}^{+}$ through upsampling $\mathcal F$ to $\mathbf H_{\mathrm g}^{+}$'s spatial resolution of $128\times128$.

To incorporate measurement-noise information into
the DiT, we employ a dedicated conditioning path for
the measurement noise level $\sigma_{\mathrm{obs}}$.
Its embedding is computed as
\begin{equation}
\mathbf e_{\sigma}
= \mathrm{MLP}_{\sigma}
\left(\frac{\sigma_{\mathrm{obs}}}{\sigma_{\mathrm{ref}}}\right),
\label{eq:noise_conditioning}
\end{equation}
where $\sigma_{\mathrm{ref}}$ is a fixed
normalization constant and $\mathrm{MLP}_{\sigma}$
is a learnable embedding network.

The measurement noise-level embedding $\mathbf e_{\sigma}$
is injected into the observation cross-attention
module by replacing its layer normalization
with noise-conditioned adaptive layer normalization:
\begin{equation}
\begin{aligned}
\widehat{\mathbf H}_{\mathrm{g}}^{+}
&=\operatorname{AdaLN}
(\mathbf H_{\mathrm{g}}^{+},\mathbf e_{\sigma})\\
&=\left(1+\gamma(\mathbf e_{\sigma})\right)
\odot\operatorname{LN}(\mathbf H_{\mathrm{g}}^{+})
+\beta(\mathbf e_{\sigma}),
\end{aligned}
\label{eq:adaptive_normalization}
\end{equation}
where $\operatorname{LN}(\cdot)$ denotes layer
normalization, $\widehat{\mathbf H}_{\mathrm{g}}^{+}$ denotes
the generation features used for the cross-attention operator 
\texttt{CrossAttn} of observations, and $\gamma$ and $\beta$
are learnable projections for feature-wise
scaling and shifting. The noise-level embedding
further controls the gating of observation
residuals injected into the generation tokens:
\begin{align}
&\mathbf H_{\mathrm{out}}\label{eq:observation_fusion}
\\
=&\mathbf H_{\mathrm{g}}^{+}
+g_{\sigma}(\mathbf e_{\sigma})\odot\operatorname{CrossAttn}\big(
\mathbf Q(\widehat{\mathbf H}_{\mathrm{g}}^{+}),
\mathbf K(\mathbf H_{\mathrm{obs}}),
\mathbf V(\mathbf H_{\mathrm{obs}})
\big),\nonumber
\end{align}
where $g_{\sigma}(\cdot)$ denotes a learnable gating function,
and $\mathbf Q(\cdot)$, $\mathbf K(\cdot)$, and
$\mathbf V(\cdot)$ denote the query, key, and value
projections, respectively.
The noise-conditioned generation features
$\widehat{\mathbf H}_{\mathrm g}^{+}$ provide queries,
while the observation features $\mathbf H_{\mathrm{obs}}$
provide keys and values, enabling generation tokens to
retrieve observation information relevant to the current
generation state.
Meanwhile, the noise-level embedding $\mathbf e_{\sigma}$
modulates the query features, and
$g_{\sigma}(\mathbf e_{\sigma})$ gates the resulting
observation residual before fusion.

The fused generation features $\mathbf H_{\mathrm{out}}$
are then processed by a stack of $N$ DiT blocks, producing
the final predicted diffusion noise $\hat{\boldsymbol{\epsilon}}$ by spatiotemporal self-attention and
a feed-forward network (FFN).
To adapt to different diffusion steps,
each DiT block encodes the diffusion timestep $\tau$ into an embedding
$\mathbf e_\tau$ as follows:
\begin{equation}
\mathbf e_\tau
= \mathrm{MLP}_{\tau}(\mathrm{PE}(\tau)),
\label{eq:timestep_embedding}
\end{equation}
where $\mathrm{PE}(\cdot)$ denotes the diffusion-step
encoding. Using the same mechanism as the noise-conditioned AdaLN in
Eq. (\ref{eq:adaptive_normalization}), $\mathbf e_\tau$ modulates
the adaptive layer normalization in each DiT block according
to the current denoising step.

Following the noise-prediction objective in Eq. (\ref{eq:diffusion_loss}), we use the same loss, with the factor $\frac{1}{LHW}$ normalizing the prediction error over all spatial locations and frames:
\begin{equation}
\mathcal L_{\mathrm{diff}}
=
\mathbb E_{\tau,\boldsymbol{\epsilon}}
\left[\frac{1}{LHW}
\|\boldsymbol{\epsilon}
-\hat{\boldsymbol{\epsilon}}\|_2^2
\right].
\label{eq:stage2_diff_loss}
\end{equation}

To improve reconstruction accuracy at sampled
locations, we introduce an observation alignment
loss on the predicted clean RM sequence.
According to Eq. (\ref{eq:ddim_sampling}), the clean RM estimate at diffusion step $\tau$
is obtained as
\begin{equation}
\hat{\mathbf X}_{1:L}^{(0)}
=
\frac{
\mathbf X_{1:L}^{(\tau)}
-\sqrt{1-\bar{\alpha}_\tau}\hat{\boldsymbol{\epsilon}}
}{
\sqrt{\bar{\alpha}_\tau}
}.
\label{eq:clean_rm_prediction}
\end{equation}
The observation alignment loss is then defined as
\begin{equation}
\begin{aligned}
\mathcal L_{\mathrm{align}}
=\mathbb E_{\tau,\boldsymbol{\epsilon}}
\Bigg[
\frac{1}{N_{\mathrm{sam}}}
\sum_{l=1}^{L}
\sum_{i\in\Omega_{\mathrm{sam},l}}
\left(\hat{x}_{l,i}^{(0)}-x_{l,i}\right)^2
\Bigg],
\end{aligned}
\label{eq:observation_alignment}
\end{equation}
where $\Omega_{\mathrm{sam},l}$ denotes the set of
sampled locations in frame $l$, and
$N_{\mathrm{sam}}=\sum_{l=1}^{L}|\Omega_{\mathrm{sam},l}|$
is the total number of sampled locations.
The alignment loss uses clean ground-truth RM
values as supervision to avoid directly fitting
measurement noise.

Both stages are jointly optimized using the following training objective:
\begin{equation}
\begin{aligned}
\mathcal L_{\mathrm{total}}
= \lambda_{\mathrm{aux}}\mathcal L_{\mathrm{aux}}
+\lambda_{\mathrm{diff}}\mathcal L_{\mathrm{diff}}
+\lambda_{\mathrm{align}}\mathcal L_{\mathrm{align}},
\end{aligned}
\label{eq:total_training_loss}
\end{equation}
where $\lambda_{\mathrm{aux}}$, $\lambda_{\mathrm{diff}}$, and $\lambda_{\mathrm{align}}$ are the loss weights. The complete training procedure is summarized in Algorithm~\ref{alg:training}.

\begin{algorithm}[!t]
\caption{Training of the Noise-Conditioned Spatiotemporal DM in DyRadio}
\label{alg:training}
\begin{algorithmic}[1]
\Require Clean RM sequences $\mathbf X_{1:L}$, scene information
$\mathbf C_{1:L}$, sparse RM observations $\mathcal Y_{1:L}$,
measurement noise level $\sigma_{\mathrm{obs}}$
\Ensure Trained U-Net $U_\phi$ and DiT $D_\theta$

\State Initialize parameters $\phi$ and $\theta$
\While{not converged}
    \State Select a mini-batch of input sequences

    \State \textbf{Stage I: Propagation-Guided Feature Extraction}
    \State $(\widetilde{\mathbf X}_{1:L},\mathcal F)
    \gets U_\phi(\mathbf C_{1:L},\mathcal Y_{1:L})$
    \State Compute $\mathcal L_{\mathrm{MSE}}$ using
    Eq.~\eqref{eq:prior_mse}
    \State Compute $\mathcal L_{\mathrm{occ}}$ and
    $\mathcal L_{\mathrm{src}}$ using
    Eqs.~\eqref{eq:prior_obstacle} and
    \eqref{eq:prior_source}
    \State Compute $\mathcal L_{\mathrm{aux}}$ using
    Eq.~\eqref{eq:prior_total}

    \State \textbf{Stage II: Noise-Conditioned Diffusion}
    \State Sample a diffusion step $\tau$
    \State $\boldsymbol{\epsilon}
    \sim \mathcal N(\mathbf 0,\mathbf I)$
    \State $\mathbf X_{1:L}^{(\tau)}
    \gets \sqrt{\bar{\alpha}_\tau}\mathbf X_{1:L}
    + \sqrt{1-\bar{\alpha}_\tau}\boldsymbol{\epsilon}$
    \State $\hat{\boldsymbol{\epsilon}}
    \gets D_\theta(
        \mathbf X_{1:L}^{(\tau)},\tau,\mathbf C_{1:L},
        \mathcal Y_{1:L},\sigma_{\mathrm{obs}},\mathcal F
    )$
    \State Compute $\mathcal L_{\mathrm{diff}}$ and
    $\mathcal L_{\mathrm{align}}$ using
    Eqs.~\eqref{eq:stage2_diff_loss} and
    \eqref{eq:observation_alignment}
    \State Compute $\mathcal L_{\mathrm{total}}$ using
    Eq.~\eqref{eq:total_training_loss}
    \State Update $\phi$ and $\theta$ using
    $\nabla_{\phi,\theta}\mathcal L_{\mathrm{total}}$
\EndWhile
\State \Return $U_\phi,D_\theta$
\end{algorithmic}
\end{algorithm}

\subsection{Iterative Noise Estimation via Blind-Area Predictive Likelihood}
\label{sec:noise-estimation}


\subsubsection{Blind-area prediction with the learned RM prior}

Consistent with Eq. \eqref{eq:observation_model}, the measurement noise is independent of the clean RM and
mutually independent across observed locations, with a common
standard deviation $\sigma_{\mathrm{obs}}$ throughout the
sequence. Therefore, we use blind-area prediction to separate
the observations used to condition each prediction from the
measurements used to evaluate its residuals. 
Specifically, we partition the observed locations $\Omega_{\mathrm{sam}}$ into $K$ disjoint blind areas:
\begin{equation}
    \Omega_{\mathrm{sam}}
    =
    \bigcup_{k=1}^{K}\Omega_{\mathrm{sam}}^{(k)},
    \qquad
    \Omega_{\mathrm{sam}}^{(i)}
    \cap
    \Omega_{\mathrm{sam}}^{(j)}
    =
    \varnothing,
    \quad i\neq j.
    \label{eq:blind-area-partition}
\end{equation}

For each spatiotemporal location
$\omega\in\Omega_{\mathrm{sam}}$, let $x_\omega$ and
$y_\omega$ denote the scalar values at location $\omega$
in the clean RM sequence $\mathbf X$ and the noisy
observation collection $\mathcal Y$, respectively. The observations retained after excluding the
$k$-th blind area are denoted by
\begin{equation}
    \mathcal{Y}_{-\Omega_{\mathrm{sam}}^{(k)}}
    =
    \left\{
        y_\omega :
        \omega\in
        \Omega_{\mathrm{sam}}
        \setminus
        \Omega_{\mathrm{sam}}^{(k)}
    \right\}.
    \label{eq:blind-area-observations}
\end{equation}
The predictive distribution depends on the current noise
estimate supplied to the DM, while the noise estimate itself
is obtained from the predictive likelihood of blind-area
measurements. This coupling motivates an iterative procedure
that alternates between generating blind-area predictions
under the current noise estimate and updating the noise level
through likelihood optimization. The updated estimate is
then used to condition the predictions in the next iteration.

At iteration $r$, the pretrained DM takes the scene information $\mathbf C$, the observations $\mathcal{Y}_{-\Omega_{\mathrm{sam}}^{(k)}}$, and the current noise estimate $\sigma_{\mathrm{obs}}^{(r)}$ as inputs, yielding the predictive distribution
\begin{equation}
    p_{\phi,\theta}
    \left(
        \mathbf X_{1:L}
        \mid
        \mathbf C,
        \mathcal{Y}_{-\Omega_{\mathrm{sam}}^{(k)}},
        \sigma_{\mathrm{obs}}^{(r)}
    \right).
    \label{eq:conditional-rm-distribution}
\end{equation}
Afterward, we generate
$S$ independent RM sequences for each blind area:
\begin{equation}
\begin{aligned}
    \widehat{\mathbf X}_{1:L}^{(r,k,s)}
    \sim\;&
    p_{\phi,\theta}
    \left(
        \widehat{\mathbf X}_{1:L}
        \mid
        \mathbf C,
        \mathcal{Y}_{-\Omega_{\mathrm{sam}}^{(k)}},
        \sigma_{\mathrm{obs}}^{(r)}
    \right),\\
    &s=1,\ldots,S.
\end{aligned}
\label{eq:monte-carlo-rm-samples}
\end{equation}
For each $\omega\in\Omega_{\mathrm{sam}}^{(k)}$,
let $\hat{x}_\omega^{(r,k,s)}$ denote the RM value at location
$\omega$ in the $s$-th generated sequence
$\widehat{\mathbf X}_{1:L}^{(r,k,s)}$.
For each $\omega\in\Omega_{\mathrm{sam}}^{(k)}$, we
approximate the marginal distribution of the clean RM value
by a Gaussian:
\begin{equation}
    p_{\phi,\theta}
    \left(
        x_\omega
        \mid
        \mathbf C,
        \mathcal{Y}_{-\Omega_{\mathrm{sam}}^{(k)}},
        \sigma_{\mathrm{obs}}^{(r)}
    \right)
    \approx
    \mathcal{N}
    \left(
        x_\omega;
        \hat{\mu}_\omega^{(r,k)},
        \hat{v}_\omega^{(r,k)}
    \right),
    \label{eq:gaussian-predictive-marginal}
\end{equation}
where the predictive mean and variance can be estimated using Monte Carlo sampling as
\begin{equation}
\begin{aligned}
    \hat{\mu}_\omega^{(r,k)}
    &=
    \frac{1}{S}
    \sum_{s=1}^{S}
    \hat{x}_\omega^{(r,k,s)},\\[4pt]
    \hat{v}_\omega^{(r,k)}
    &=
    \frac{1}{S-1}
    \sum_{s=1}^{S}
    \left(
        \hat{x}_\omega^{(r,k,s)}
        -
        \hat{\mu}_\omega^{(r,k)}
    \right)^2.
\end{aligned}
\label{eq:monte-carlo-predictive-moments}
\end{equation}

\subsubsection{Noise estimation through predictive likelihood}

We obtain
the predictive density of the blind-area measurement by
marginalizing over the clean RM value:
    \begin{align}
        &p_\omega^{(r,k)}(y_\omega;\sigma_{\mathrm{obs}})\nonumber\\
        &=
        \int
        p(y_\omega\mid x_\omega,\sigma_{\mathrm{obs}})
        p_{\phi,\theta}
        \left(
            x_\omega
            \mid
            \mathbf C,
            \mathcal{Y}_{-\Omega_{\mathrm{sam}}^{(k)}},
            \sigma_{\mathrm{obs}}^{(r)}
        \right)
        \,\mathrm{d}x_\omega \nonumber\\
        &\approx
        \int
        \mathcal{N}
        \left(
            y_\omega;x_\omega,\sigma_{\mathrm{obs}}^2
        \right)
        \mathcal{N}
        \left(
            x_\omega;
            \hat{\mu}_\omega^{(r,k)},
            \hat{v}_\omega^{(r,k)}
        \right)
        \,\mathrm{d}x_\omega    \label{eq:measurement-predictive-density}
\\
        &=
        \mathcal{N}
        \left(
            y_\omega;
            \hat{\mu}_\omega^{(r,k)},
            \hat{v}_\omega^{(r,k)}
            +\sigma_{\mathrm{obs}}^2
        \right).\nonumber
    \end{align}
Here, $\sigma_{\mathrm{obs}}$ is the noise standard
deviation being estimated. The predictive variance is the sum of the clean RM prediction
variance $\hat{v}_\omega^{(r,k)}$ and the measurement-noise
variance $\sigma_{\mathrm{obs}}^2$. Consequently, the
predictive likelihood enables noise estimation from blind-area
measurements while accounting for uncertainty in the clean RM
prediction.

We combine the Gaussian predictive densities in
Eq.~(\ref{eq:measurement-predictive-density}) across all
blind areas to form the composite predictive likelihood:
\begin{equation}
    \mathcal{P}^{(r)}(\sigma_{\mathrm{obs}})
    \triangleq
    \prod_{k=1}^{K}
    \prod_{\omega\in\Omega_{\mathrm{sam}}^{(k)}}
    \mathcal{N}
    \left(
        y_\omega;
        \hat{\mu}_\omega^{(r,k)},
        \hat{v}_\omega^{(r,k)}
        +\sigma_{\mathrm{obs}}^2
    \right).
    \label{eq:composite-predictive-likelihood}
\end{equation}

Using maximum composite likelihood estimation, we estimate
the shared noise standard deviation by maximizing
$\mathcal{P}^{(r)}(\sigma_{\mathrm{obs}})$, that is, 
    \begin{align}
    &\widetilde{\sigma}_{\mathrm{obs}}^{(r+1)}
    =
    \underset{\sigma_{\mathrm{obs}}\geq 0}{\arg\max}\,
    \mathcal{P}^{(r)}(\sigma_{\mathrm{obs}})\nonumber\\
    =&
    \underset{\sigma_{\mathrm{obs}}\geq 0}{\arg\min}\,
    \left[
        -\log\mathcal{P}^{(r)}(\sigma_{\mathrm{obs}})
    \right]\label{eq:noise-estimation-objective}\\
    =&
    \underset{\sigma_{\mathrm{obs}}\geq 0}{\arg\min}
    \sum_{k=1}^{K}
    \sum_{\omega\in\Omega_{\mathrm{sam}}^{(k)}}
    \Bigg[
        \log\!\left(
            \hat{v}_\omega^{(r,k)}
            +\sigma_{\mathrm{obs}}^2
        \right)
        +
        \frac{
            (y_\omega-\hat{\mu}_\omega^{(r,k)})^2
        }{
            \hat{v}_\omega^{(r,k)}
            +\sigma_{\mathrm{obs}}^2
        }
    \Bigg].\nonumber
\end{align}
The logarithmic term $\log\!\left(
            \hat{v}_\omega^{(r,k)}
            +\sigma_{\mathrm{obs}}^2
        \right)$ penalizes excessive predictive variance,
while the normalized residual term $\frac{
            (y_\omega-\hat{\mu}_\omega^{(r,k)})^2
        }{
            \hat{v}_\omega^{(r,k)}
            +\sigma_{\mathrm{obs}}^2
        }$ penalizes variance that
is too small to explain the observed measurements. Together,
these terms estimate the shared noise level while accounting
for location-dependent predictive uncertainty.

We solve the optimization problem numerically and apply
relaxation to update the noise estimate:
\begin{equation}
    \sigma_{\mathrm{obs}}^{(r+1)}
    =
    (1-\eta)\sigma_{\mathrm{obs}}^{(r)}
    +
    \eta\widetilde{\sigma}_{\mathrm{obs}}^{(r+1)},
    \label{eq:damped-noise-update}
\end{equation}
where $\eta\in(0,1]$ is the relaxation parameter. 
Starting from an initial estimate
$\sigma_{\mathrm{obs}}^{(0)}$, we repeat the prediction 
and update steps in Eq. (\ref{eq:monte-carlo-rm-samples}) and Eq. (\ref{eq:damped-noise-update}) respectively until $\left|
        \sigma_{\mathrm{obs}}^{(r+1)}
        -\sigma_{\mathrm{obs}}^{(r)}
    \right|<\delta$ 
or the maximum number of iterations $R_{\max}$ is reached, yielding the final estimate $\hat{\sigma}_{\mathrm{obs}}$.

By Eq.~\eqref{eq:dit_prediction}, the final estimate $\hat{\sigma}_{\mathrm{obs}}$, together
with the scene information $\mathbf C$ and all available
measurements $\mathcal{Y}_{1:T}$, conditions the pretrained
spatiotemporal DM to reconstruct the complete RM sequence.
Algorithm~\ref{alg:noise-estimation} summarizes the
estimation and reconstruction procedure of DyRadio.

\begin{algorithm}[!t]
\caption{DyRadio: Iterative Noise Estimation and Dynamic RM Reconstruction}
\label{alg:noise-estimation}
\begin{algorithmic}[1]
\Require Pretrained $U_\phi$ and $D_\theta$,
scene information $\mathbf C$,
sparse observations $\mathcal Y_{1:T}$,
initial noise estimate $\sigma_{\mathrm{obs}}^{(0)}$
\Ensure Noise estimate $\hat{\sigma}_{\mathrm{obs}}$
and reconstructed RM sequence $\widehat{\mathbf X}_{1:T}$

\Statex \textbf{Noise Estimation}
\State Select an initial subset of observations for noise
estimation, with sampled locations $\Omega_{\mathrm{sam}}$
\State Partition $\Omega_{\mathrm{sam}}$ into $K$ disjoint blind areas
$\{\Omega_{\mathrm{sam}}^{(k)}\}_{k=1}^{K}$
\State $r \gets 0$

\Repeat
    \For{$k=1,\ldots,K$}
        \State Form $\mathcal Y_{-\Omega_{\mathrm{sam}}^{(k)}}$ using Eq.~\eqref{eq:blind-area-observations}
        \State Draw $S$ independent RM sequences using
        Eq. (\ref{eq:monte-carlo-rm-samples})
        \State Estimate
        $\{\hat{\mu}_\omega^{(r,k)},
        \hat{v}_\omega^{(r,k)}\}_{\omega\in\Omega_{\mathrm{sam}}^{(k)}}$
        using Eq. (\ref{eq:monte-carlo-predictive-moments})
    \EndFor

    \State Compute $\widetilde{\sigma}_{\mathrm{obs}}^{(r+1)}$
    by solving Eq. (\ref{eq:noise-estimation-objective})

    \State Update $\sigma_{\mathrm{obs}}^{(r+1)}
    \gets (1-\eta)\sigma_{\mathrm{obs}}^{(r)}
    +\eta\widetilde{\sigma}_{\mathrm{obs}}^{(r+1)}$ as Eq.~\eqref{eq:damped-noise-update}  
    \State $\Delta
    \gets \left|
        \sigma_{\mathrm{obs}}^{(r+1)}
        -\sigma_{\mathrm{obs}}^{(r)}
    \right|$
    \State $r \gets r+1$
\Until{$\Delta<\delta$
    \textbf{or} $r\geq R_{\max}$}

\State $\hat{\sigma}_{\mathrm{obs}}
\gets \sigma_{\mathrm{obs}}^{(r)}$

\Statex \textbf{RM Reconstruction}
\State Reconstruct all $L$-frame subsequences using
$U_\phi$ and $D_\theta$, conditioned on the
corresponding scene information $\mathbf C$, sparse observations,
and $\hat{\sigma}_{\mathrm{obs}}$ using Eq.~\eqref{eq:auxiliary_prediction} and Eq.~\eqref{eq:dit_prediction}
\State \Return $\hat{\sigma}_{\mathrm{obs}},
\widehat{\mathbf X}_{1:T}$
\end{algorithmic}
\end{algorithm}


\section{Experimental Settings and Results}
\label{sec:experiments}
In this section, we evaluate the performance of DyRadio for dynamic radio map reconstruction and compare DyRadio against various baseline methods, such as RadioDiff~\cite{wang2025radiodiff}, PhyRMDM~\cite{jia2025physics}, RadioUNet~\cite{levie2021radiounet}, and RadioMapMotion~\cite{cheng2026radiomapmotion}.
\subsection{Experimental Settings}
\begin{table}[!t]
    \centering
    \caption{Key Model Configurations and Training Settings of DyRadio.}
    \label{tab:model_config}
    \renewcommand{\arraystretch}{1.08}
    \setlength{\tabcolsep}{3pt}
    \renewcommand{\tabularxcolumn}[1]{m{#1}}
    \begin{tabularx}{1\columnwidth}{
        @{}
        >{\raggedright\arraybackslash}m{0.44\columnwidth}
        >{\raggedright\arraybackslash}X
        @{}
    }
        \toprule
        Parameter & Configuration \\
        \midrule

        3D patch size
        & $(2,4,4)$ \\

        Token grid size
        & $(8,32,32)$ \\

        U-Net base channels
        & $32$ \\

        U-Net down/up stages
        & $5/5$ \\

        DiT block counts $N$
        & $2/11/2$ (first/middle/last) \\

        Feature dimensions
        & $384/768/384$ (first/middle/last) \\

        Attention heads
        & $6/12/6$ (first/middle/last) \\

        $p$ range for training
        & $1\%\text{--}10\%$ \\

        $\sigma_{\mathrm{obs}}$ range for training
        & $0.01\text{--}0.09$ \\

        Learning rate
        & $5\times10^{-5}$ \\

        Weight decay
        & $0.01$ \\

        Batch size
        & $24$ \\

        Learning rate schedule
        & Cosine decay with 1000 warmup steps \\

        Noise estimation frames
        & First $4$ frames of each sequence \\
        
        Initial noise estimate $\sigma_{\mathrm{obs}}^{(0)}$
        & $\sigma_{\mathrm{obs}}^{(0)}=0.03$ \\

        Relaxation parameter $\eta$
        & $\eta=0.8$ \\

        Maximum number of iterations $R_{\max}$
        & $R_{\max}=3$ \\
        Number of blind areas $K$
        & $K=4$ \\

        Monte Carlo samples $S$
        & $S=4$ \\

        Convergence tolerance $\delta$
        & $\delta=0.001+0.02\sigma_{\mathrm{obs}}^{(r)}$ \\

        \bottomrule
    \end{tabularx}
\end{table}
\begin{figure*}[t]
    \centering
    \includegraphics[width=\textwidth]{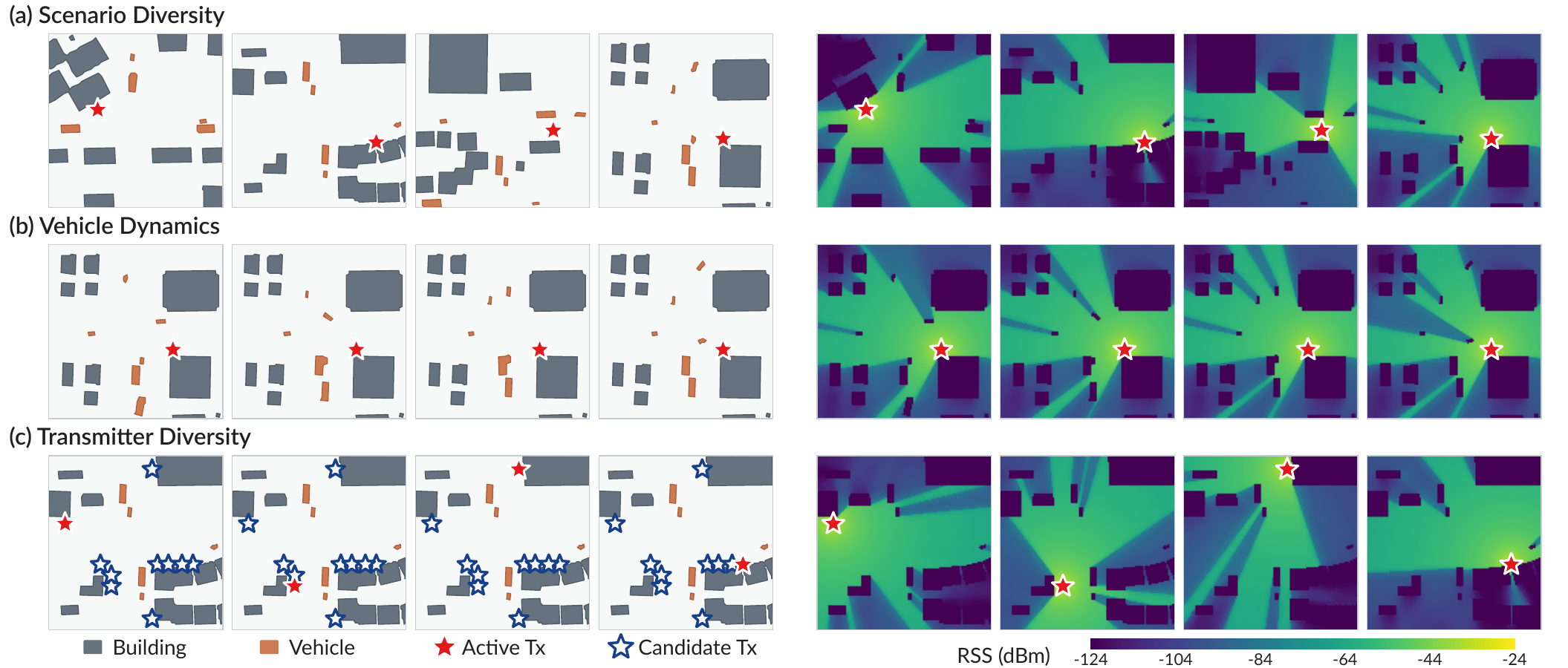}
    \caption{Visualization of scenario diversity, vehicle dynamics,
and transmitter diversity in the generated dataset.}
    \label{fig:dataset_visualization}
\end{figure*}
\begin{table*}[!t]
\centering
\caption{Quantitative Comparison of RM Reconstruction Methods.}
\label{tab:main_results}
\footnotesize
\setlength{\tabcolsep}{2.5pt}
\renewcommand{\arraystretch}{1.05}
\begin{tabular*}{\textwidth}{@{\extracolsep{\fill}}ll*{12}{c}@{}}
\toprule
\multirow{2}{*}{$p$} &
\multirow{2}{*}{Method} &
\multicolumn{4}{c}{$\sigma_{\mathrm{obs}}=0.03$} &
\multicolumn{4}{c}{$\sigma_{\mathrm{obs}}=0.05$} &
\multicolumn{4}{c}{$\sigma_{\mathrm{obs}}=0.09$} \\
\cline{3-6}\cline{7-10}\cline{11-14}
& & MSE$\downarrow$ & NMSE$\downarrow$ & PSNR$\uparrow$ & SSIM$\uparrow$
& MSE$\downarrow$ & NMSE$\downarrow$ & PSNR$\uparrow$ & SSIM$\uparrow$
& MSE$\downarrow$ & NMSE$\downarrow$ & PSNR$\uparrow$ & SSIM$\uparrow$ \\
\hline

\multirow{5}{*}{1\%}
& RadioDiff
& 1.6920 & 6.497 & 27.97 & 0.9444
& 1.9893 & 7.634 & 27.24 & 0.9375
& 3.0045 & 11.512 & 25.42 & \underline{0.9180} \\

& PhyRMDM
& 1.7204 & 6.638 & 27.84 & 0.8205
& 1.9881 & 7.664 & 27.19 & 0.8149
& 3.0093 & 11.560 & 25.35 & 0.7964 \\

& RadioUNet
& 2.0286 & 7.783 & 27.10 & 0.9338
& 2.3068 & 8.833 & 26.51 & 0.9252
& 3.3170 & 12.646 & 24.88 & 0.8968 \\

& RadioMapMotion
& \underline{0.7665} & \underline{2.918} & \underline{31.28} & \underline{0.9613}
& \underline{1.0540} & \underline{4.014} & \underline{29.85} & \underline{0.9421}
& \underline{2.8323} & \underline{10.714} & \underline{25.50} & 0.8486 \\

& \textbf{DyRadio}
& \textbf{0.5355} & \textbf{2.037} & \textbf{32.81} & \textbf{0.9777}
& \textbf{0.5911} & \textbf{2.250} & \textbf{32.38} & \textbf{0.9757}
& \textbf{0.7680} & \textbf{2.928} & \textbf{31.25} & \textbf{0.9696} \\

\hline

\multirow{5}{*}{2\%}
& RadioDiff
& 1.2193 & 4.670 & 29.35 & 0.9562
& 1.3623 & 5.217 & 28.85 & 0.9519
& \underline{1.8196} & \underline{6.968} & \underline{27.57} & \underline{0.9400} \\

& PhyRMDM
& 1.2433 & 4.783 & 29.20 & 0.8220
& 1.3905 & 5.350 & 28.71 & 0.8184
& 1.9385 & 7.454 & 27.24 & 0.8055 \\

& RadioUNet
& 1.3211 & 5.063 & 28.94 & 0.9502
& 1.4975 & 5.730 & 28.37 & 0.9431
& 2.1547 & 8.213 & 26.74 & 0.9177 \\

& RadioMapMotion
& \underline{0.4776} & \underline{1.819} & \underline{33.36} & \underline{0.9723}
& \underline{0.6531} & \underline{2.488} & \underline{31.94} & \underline{0.9578}
& 1.8543 & 7.030 & 27.34 & 0.8787 \\

& \textbf{DyRadio}
& \textbf{0.3478} & \textbf{1.323} & \textbf{34.67} & \textbf{0.9845}
& \textbf{0.3838} & \textbf{1.462} & \textbf{34.24} & \textbf{0.9830}
& \textbf{0.4952} & \textbf{1.890} & \textbf{33.14} & \textbf{0.9785} \\

\hline

\multirow{5}{*}{3\%}
& RadioDiff
& 1.0361 & 3.964 & 30.03 & \underline{0.9615}
& 1.1295 & 4.321 & 29.65 & \underline{0.9584}
& 1.4233 & 5.445 & 28.62 & \underline{0.9495} \\

& PhyRMDM
& 1.1174 & 4.292 & 29.64 & 0.8143
& 1.2285 & 4.720 & 29.22 & 0.8116
& 1.6322 & 6.273 & 27.97 & 0.8011 \\

& RadioUNet
& 1.0360 & 3.967 & 29.99 & 0.9573
& 1.1704 & 4.477 & 29.44 & 0.9509
& 1.6771 & 6.394 & 27.83 & 0.9275 \\

& RadioMapMotion
& \underline{0.3715} & \underline{1.415} & \underline{34.45} & \underline{0.9766}
& \underline{0.4971} & \underline{1.893} & \underline{33.14} & \underline{0.9648}
& \underline{1.3391} & \underline{5.085} & \underline{28.76} & 0.9002 \\

& \textbf{DyRadio}
& \textbf{0.2781} & \textbf{1.058} & \textbf{35.63} & \textbf{0.9871}
& \textbf{0.3066} & \textbf{1.168} & \textbf{35.21} & \textbf{0.9858}
& \textbf{0.3949} & \textbf{1.507} & \textbf{34.12} & \textbf{0.9819} \\

\bottomrule
\end{tabular*}

\par\vspace{2pt}
\parbox{\textwidth}{\footnotesize
\textit{Note:} MSE and NMSE are reported in $10^{-3}$. Best and second-best results are highlighted
in \textbf{bold} and \underline{underlined}, respectively.}
\end{table*}
\subsubsection{Dataset}
We construct a dynamic RM dataset by integrating the CARLA
simulator~\cite{dosovitskiy2017carla} with Sionna RT~\cite{hoydis2023sionna}. Specifically, CARLA is employed to
construct 3D environments and simulate realistic vehicle
movements, generating diverse traffic trajectories. The scene
geometry and dynamic vehicle configurations are then imported
into Sionna RT to perform ray-tracing-based radio propagation
simulations. The Sionna RT simulations account for multipath propagation
through surface reflections and Lambertian diffuse scattering
from buildings, roads, and vehicles. The carrier frequency is set to $5.89$ GHz, with a
bandwidth of $10$ MHz and a transmit power of $23$ dBm. The transmitter
and receiver heights are set to $1.5$ m and $1.0$ m, respectively. 

The dataset consists of $16$ different scenes, each containing
$150$ episodes with distinct vehicle trajectories. For each scene,
$40$ candidate transmitter locations are randomly deployed and $5$ are selected per episode to generate $5$
RM sequences, 
resulting in a total of $12,000$ RM sequences. 
Each RM represents the spatial distribution of received signal
strength (RSS), normalized to $[0,1]$, on a $128\times128$ grid
covering a $96\times96$ m$^2$ area. Each sequence is generated
at $10$ Hz, with $96$ frames. Fig.~\ref{fig:dataset_visualization} illustrates the diversity
of the generated dataset, which capably captures the temporal variations in
wireless propagation caused by moving vehicles, including
changes in signal blockage and multipath propagation. The dataset is further divided into $12$ training scenes, one validation
scene, and three testing scenes, with no scene overlap. To keep the computational cost manageable, we randomly select
$1,800$, $120$, and $300$
sequences for training, validation, and testing,
respectively, while each training sequence is divided into six
non-overlapping $L=16$-frame subsequences. During evaluation, we reconstruct complete 96-frame
sequences to assess the performance of DyRadio in dynamic
environments.

\begin{figure*}[!t]
    \centering
    \includegraphics[width=1\textwidth]{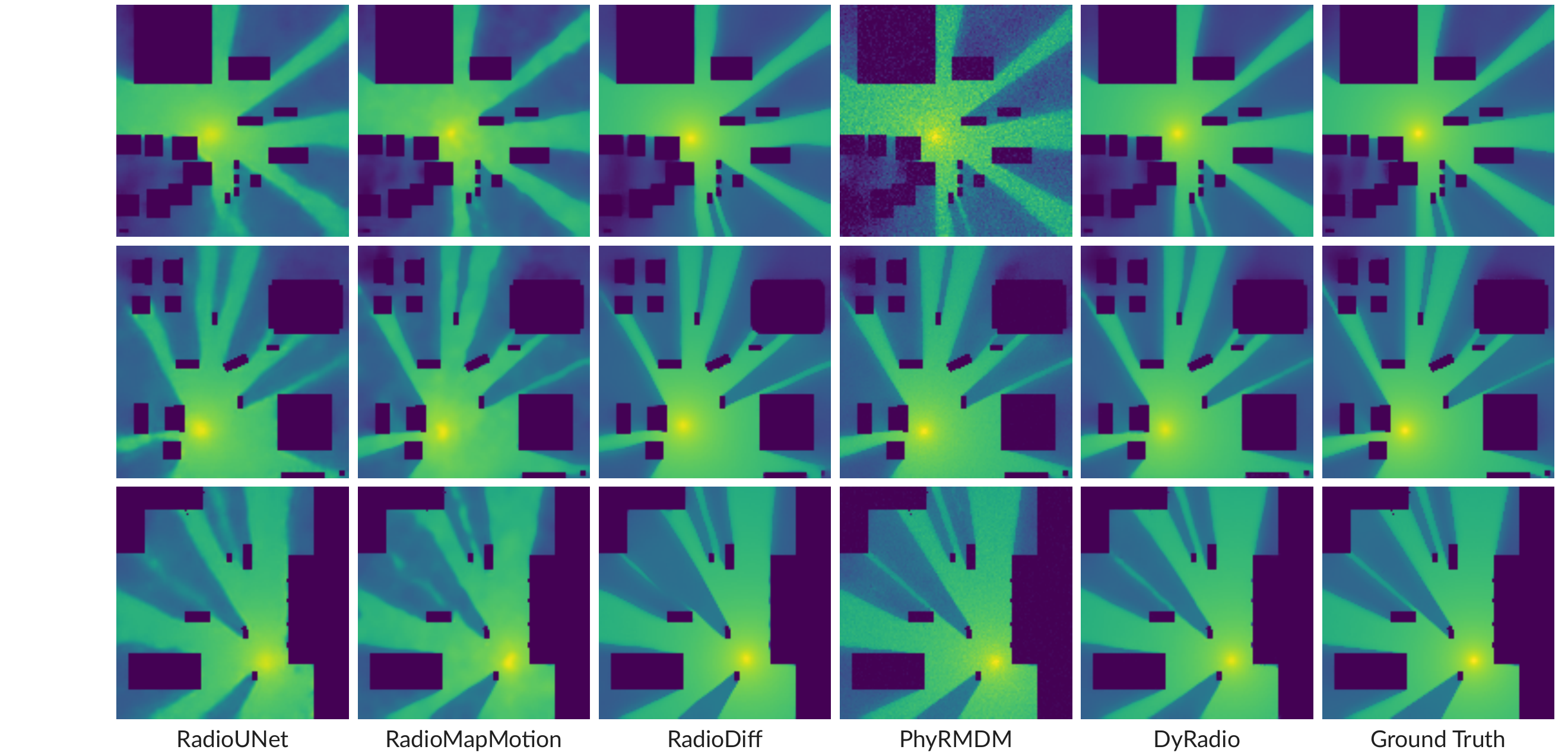}
    \vspace{-12pt}
    \caption{Qualitative comparison of RM reconstruction results at different time instants (\(p=2\%\), \(\sigma_{\mathrm{obs}}=0.05\)).}
    \label{fig:generation_results}
\end{figure*}

\subsubsection{Implementation Details}



To balance reconstruction performance and computational cost,
the DiT blocks are divided into three groups: first, middle,
and last. Different feature dimensions and numbers of attention
heads are assigned to each group. Stage I and Stage II are jointly trained from scratch using
AdamW on training sequences with varying sampling rates and
observation noise levels, as specified in
Table~\ref{tab:model_config}. For noise estimation, each blind
area is evaluated using 10-step DDIM sampling, while RM
reconstruction uses 50-step DDIM sampling. All models are trained on four NVIDIA RTX PRO 6000 GPUs, and the key model configurations and
training settings are summarized in Table~\ref{tab:model_config}.

\subsubsection{Comparison Methods}
We compare DyRadio with four representative RM reconstruction
methods covering both single- and multi-frame settings, as well
as diffusion-based and supervised approaches.
\begin{itemize}
\item \textbf{RadioDiff~\cite{wang2025radiodiff}:}
RadioDiff is a sampling-free, single-frame diffusion-based
method for RM reconstruction. It employs latent-space diffusion with an
attention-based U-Net and adaptive Fourier transform modules
to capture propagation features, together with a decoupled
diffusion process for efficient generation.

\item \textbf{PhyRMDM~\cite{jia2025physics}:}
PhyRMDM is a single-frame, physics-informed diffusion model
for sparse RM reconstruction. It adopts a dual U-Net architecture that
incorporates Helmholtz equation constraints into
diffusion-based generation through a representation
alignment mechanism.

\item \textbf{RadioUNet~\cite{levie2021radiounet}:}
RadioUNet is a supervised, single-frame CNN-based method for
direct RM prediction. Its U-Net learns a deterministic mapping
from environmental information to spatial radio fields and
supports sparse-measurement-assisted reconstruction.

\item \textbf{RadioMapMotion~\cite{cheng2026radiomapmotion}:}
RadioMapMotion is a supervised, multi-frame direct-prediction
model that uses ConvLSTM to fuse features across sequential
inputs. We adapt it for sparse-observation RM reconstruction
while retaining the original architecture and retraining
the model.
\end{itemize}
For a fair comparison, sparse observations are incorporated into baselines that do not originally support them. All baselines are retrained using identical dataset splits, with early stopping based on validation loss. The definitions of all used evaluation metrics are provided in the Appendix.

\subsection{Experimental Results and Analysis}
\subsubsection{Overall Reconstruction Performance}
Table~\ref{tab:main_results} compares DyRadio with four baselines
across sampling rates and noise levels. DyRadio consistently
achieves the lowest MSE and NMSE and the highest PSNR and SSIM
across all settings. Single-frame RM methods~\cite{wang2025radiodiff, jia2025physics,levie2021radiounet} lack explicit cross-frame modeling, limiting
their use of complementary observations. As shown in
Fig.~\ref{fig:generation_results}, these methods produce distorted
local propagation patterns due to insufficient observations in
the corresponding regions. Compared with RadioMapMotion~\cite{cheng2026radiomapmotion},
which uses ConvLSTM for temporal fusion, DyRadio reduces MSE
by $25.1\%$--$73.3\%$, highlighting the effectiveness of
DiT-based spatiotemporal modeling in exploiting complementary
observations. Moreover, Fig.~\ref{fig:generation_results} shows that
RadioMapMotion's direct supervision-based predictions tend to smooth
local variations, blurring shadow boundaries. These gains are particularly pronounced at high noise
levels, highlighting how noise-conditioned observation fusion
adapts sparse-observation integration to the estimated noise level
and improves robustness.

\subsubsection{Impact of Sampling Rate and Measurement Noise on RM Reconstruction Performance}

\begin{figure}[t]
    \centering
\includegraphics[width=\columnwidth]{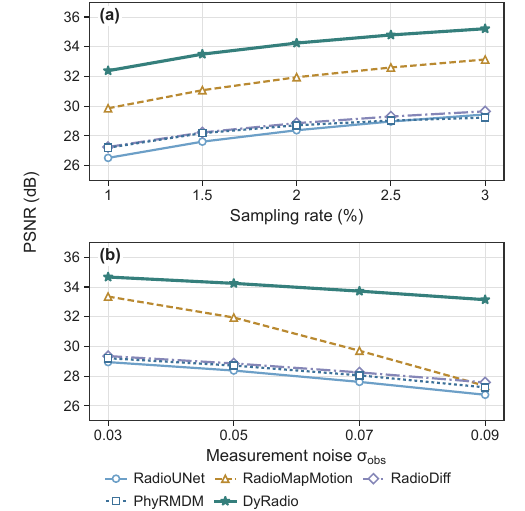}
    \vspace{-12pt}
    \caption{Impact of (a) sampling rate ($\sigma_{\mathrm{obs}}=0.05$) and (b) measurement noise ($p=2\%$) on RM reconstruction performance.}
    \label{fig:psnr_impacts}
\end{figure}

Fig.~\ref{fig:psnr_impacts} illustrates the effects of sampling rate and measurement noise on reconstruction performance. As shown in Fig.~\ref{fig:psnr_impacts}(a), the PSNR of all methods increases with the sampling rate $p$, while DyRadio consistently maintains a substantial advantage over the baselines. Specifically, its PSNR improves by $2.83$ dB as the sampling rate increases from $1\%$ to $3\%.$ Fig.~\ref{fig:psnr_impacts}(b) shows that increasing measurement noise degrades reconstruction accuracy across all methods. However, DyRadio exhibits a smaller PSNR reduction of $1.53$ dB as $\sigma_{\mathrm{obs}}$ increases from $0.03$ to $ 0.09$, whereas RadioMapMotion suffers a much larger drop of approximately $6.0$ dB, eventually approaching the performance of other baselines. This sharp degradation indicates that multi-frame fusion makes RadioMapMotion more vulnerable to noise. In contrast, DyRadio's noise-conditioned design effectively preserves reconstruction quality under increasingly noisy observations.

\begin{figure}[t]
    \centering
    \includegraphics[width=\columnwidth]
    {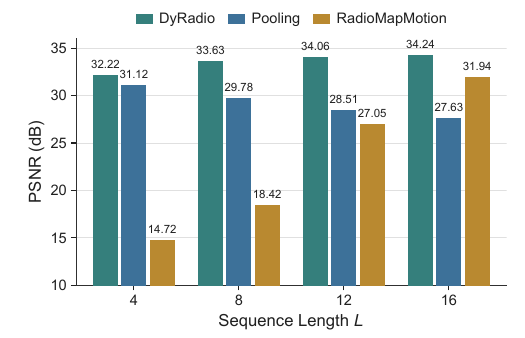}
    \vspace{-12pt}
    \caption{Effectiveness of temporal modeling for RM reconstruction ($p=2\%$, $\sigma_{\mathrm{obs}}=0.05$) under different sequence lengths.}
    \label{fig:temporal_ablation}
\end{figure}

\subsubsection{Effectiveness of Temporal Modeling}
We evaluate temporal modeling by varying the input sequence length and comparing three methods: DyRadio, a DyRadio variant that directly pools observations, and RadioMapMotion~\cite{cheng2026radiomapmotion}. Both the flexible DiT and ConvLSTM architectures support processing sequences of different lengths using the same pretrained model weights. As shown in Fig.~\ref{fig:temporal_ablation}, the direct-pooling variant of DyRadio degrades from $31.12$ to $27.63$ dB as $L$ increases from $4$ to $16$, whereas DyRadio improves from $32.22$ to $34.24$ dB, highlighting the adverse effect of temporal state mismatch. RadioMapMotion exhibits a stronger dependence on sequence length, with PSNR increasing from $14.72$ to $31.94$ dB, while DyRadio already achieves $32.22$ dB at $L=4$. These results demonstrate DyRadio's effectiveness in exploiting temporal observations while maintaining high reconstruction quality with short input sequences.

\begin{table}[t]
    \centering
    \small
    \caption{Noise Estimation Accuracy Under Different Sampling Rates and True Noise Levels.}
    \label{tab:noise_estimation}
    \setlength{\tabcolsep}{4pt}
    \renewcommand{\arraystretch}{1.05}
    \begin{tabular*}{\columnwidth}{@{\extracolsep{\fill}}ccccc@{}}
    \toprule
    \multirow{2}{*}{$p$} &
    \multirow{2}{*}{\shortstack{Noise Level\\$\sigma_{\mathrm{obs}}$}} &
    \multicolumn{3}{c}{Metric} \\
    \cline{3-5}
    & & \shortstack{Avg.\\Estimate} &
    \shortstack{MAE\\($\downarrow$)} &
    \shortstack{Relative Bias\\(\%)} \\
    \specialrule{\lightrulewidth}{1.5pt}{1.5pt}

    \multirow{5}{*}{1\%}
    & 0.00 & 0.0019 & 1.931 & -- \\
    & 0.03 & 0.0293 & 2.755 & -2.34 \\
    & 0.05 & \underline{0.0505} & 2.342 & \underline{+1.00} \\
    & 0.07 & 0.0710 & 2.806 & +1.48 \\
    & 0.09 & 0.0890 & 1.003 & -1.11 \\
    \specialrule{\lightrulewidth}{1.5pt}{1.5pt}

    \multirow{5}{*}{2\%}
    & 0.00 & \underline{0.0016} & \underline{1.642} & -- \\
    & 0.03 & \underline{0.0304} & \underline{1.365} & \underline{+1.32} \\
    & 0.05 & 0.0506 & \underline{1.464} & +1.17 \\
    & 0.07 & \underline{0.0707} & \underline{1.758} & \underline{+0.99} \\
    & 0.09 & \underline{0.0894} & \underline{0.636} & \underline{-0.71} \\
    \specialrule{\lightrulewidth}{1.5pt}{1.5pt}

    \multirow{5}{*}{3\%}
    & 0.00 & \textbf{0.0010} & \textbf{1.016} & -- \\
    & 0.03 & \textbf{0.0299} & \textbf{1.218} & \textbf{-0.38} \\
    & 0.05 & \textbf{0.0502} & \textbf{1.167} & \textbf{+0.39} \\
    & 0.07 & \textbf{0.0703} & \textbf{1.352} & \textbf{+0.41} \\
    & 0.09 & \textbf{0.0894} & \textbf{0.633} & \textbf{-0.70} \\
    \bottomrule
    \end{tabular*}

    \par\vspace{2pt}\noindent
    \parbox{\columnwidth}{\footnotesize
        \textit{Note:} MAE is reported in units of $10^{-3}$. Relative bias is reported as ``--'' when
        $\sigma_{\mathrm{obs}}=0$. Best and second-best results are highlighted in bold and underlined, respectively.
    }
    \vspace{-2mm}
\end{table}
\begin{table}[t]
    \centering
    \caption{Impact of Noise Estimation on RM Reconstruction ($p=2\%$)}
    \label{tab:noise_conditioning}
    \small
    \renewcommand{\arraystretch}{1.2}
    \begin{tabular*}{\columnwidth}{@{\extracolsep{\fill}}clcccc@{}}
    \toprule        $\sigma_{\mathrm{obs}}$ & Input $\sigma$
        & MSE & NMSE & PSNR & SSIM \\
        \hline
        \multirow{3}{*}{0.03}
        & Ground-Truth                  & 0.3476 & 1.3226 & 34.674 & 0.98454 \\
        & Estimated             & 0.3478 & 1.3234 & 34.672 & 0.98453 \\
        & Fixed ($\sigma=0.05$) & 0.3950 & 1.5030 & 34.126 & 0.98284 \\
        \midrule
        \multirow{3}{*}{0.05}
        & Ground-Truth                  & 0.3836 & 1.4609 & 34.246 & 0.98302 \\
        & Estimated             & 0.3838 & 1.4619 & 34.244 & 0.98302 \\
        & Fixed ($\sigma=0.05$) & --- & --- & --- & --- \\
        \midrule
        \multirow{3}{*}{0.07}
        & Ground-Truth                  & 0.4331 & 1.6512 & 33.722 & 0.98099 \\
        & Estimated             & 0.4333 & 1.6519 & 33.720 & 0.98099 \\
        & Fixed ($\sigma=0.05$) & 0.4488 & 1.7097 & 33.571 & 0.98063 \\
        \midrule
        \multirow{3}{*}{0.09}
        & Ground-Truth                  & 0.4953 & 1.8901 & 33.144 & 0.97854 \\
        & Estimated             & 0.4952 & 1.8899 & 33.144 & 0.97853 \\
        & Fixed ($\sigma=0.05$) & 0.5316 & 2.0388 & 32.848 & 0.97514 \\
        \bottomrule
    \end{tabular*}
    \par\vspace{2pt}
    \parbox{\columnwidth}{\footnotesize
    \textit{Note:} MSE and NMSE are reported in $10^{-3}$. Results for the Fixed ($\sigma=0.05$) are omitted, because they coincide with the ground-truth values.
    }
\end{table}

\begin{table*}[t]
    \centering
    \caption{Ablation Study of DyRadio Under Different Sampling Rates and Noise Levels.}
    \label{tab:ablation}
    \footnotesize
    \setlength{\tabcolsep}{2.5pt}
    \renewcommand{\arraystretch}{1.2}
    \begin{tabular*}{\textwidth}{@{\extracolsep{\fill}}l*{12}{c}@{}}
        \toprule
        \multirow{2}{*}{Model Variant}
        & \multicolumn{4}{c}{$p=1\%,\ \sigma_{\mathrm{obs}}=0.05$}
        & \multicolumn{4}{c}{$p=2\%,\ \sigma_{\mathrm{obs}}=0.05$}
        & \multicolumn{4}{c}{$p=2\%,\ \sigma_{\mathrm{obs}}=0.09$} \\
        \cline{2-5}\cline{6-9}\cline{10-13}
        & MSE$\downarrow$ & NMSE$\downarrow$ & PSNR$\uparrow$ & SSIM$\uparrow$
        & MSE$\downarrow$ & NMSE$\downarrow$ & PSNR$\uparrow$ & SSIM$\uparrow$
        & MSE$\downarrow$ & NMSE$\downarrow$ & PSNR$\uparrow$ & SSIM$\uparrow$ \\
        \hline
        Full Model
        & \textbf{0.591} & \textbf{2.250} & \textbf{32.38} & \textbf{0.9757}
        & \textbf{0.384} & \textbf{1.462} & \textbf{34.24} & \textbf{0.9830}
        & \textbf{0.495} & \textbf{1.890} & \textbf{33.14} & \textbf{0.9785} \\

        w/o Stage I
        & 3.932 & 14.974 & 24.10 & 0.8705
        & 1.941 & 7.400 & 27.17 & 0.9266
        & 2.759 & 10.507 & 25.63 & 0.9007 \\

        w/o Noise-Conditioned Fusion
        & 0.763 & 2.902 & 31.23 & 0.9604
        & 0.478 & 1.826 & 33.27 & 0.9712
        & 0.983 & 3.754 & 30.13 & 0.9429 \\

        w/o Temporal Attention
        & 3.244 & 12.426 & 25.05 & 0.9131
        & 1.343 & 5.127 & 28.81 & 0.9494
        & 2.042 & 7.816 & 27.02 & 0.9312 \\
        \bottomrule
    \end{tabular*}
    \vspace{2pt}

    \parbox{\textwidth}{\footnotesize
    \textit{Note:} MSE and NMSE are reported in $10^{-3}$.
    }
\end{table*}

\subsubsection{Noise Estimation Accuracy and Impact on Reconstruction}
Table~\ref{tab:noise_estimation} summarizes noise estimation accuracy under different sampling rates and true noise levels. As the sampling rate increases, the MAE consistently decreases across all noise levels, while the relative bias generally approaches zero. For example, at $\sigma_{\mathrm{obs}}=0.03$, the MAE decreases from $2.755\times10^{-3}$ to $1.218\times10^{-3}$, and the absolute relative bias drops from $2.34\%$ to $0.38\%$ as $p$ increases from $1\%$ to $3\%$. This improvement indicates that additional observations enhance the reliability of noise estimation, even under sparse sampling conditions.

At a sampling rate of $p=2\%$, Table~\ref{tab:noise_conditioning} evaluates
the impact of noise estimation on RM reconstruction
by comparing ground-truth, estimated, and fixed ($\sigma=0.05$) noise level inputs. The estimated noise levels achieve nearly identical performance to the ground-truth inputs, with a maximum PSNR difference of only $0.002$ dB. Compared with the fixed input, the estimated-noise setting improves PSNR by $0.546$, $0.149$, and $0.296$ dB at $\sigma_{\mathrm{obs}}=0.03$, $0.07$, and $0.09$, respectively. These results demonstrate that the proposed noise estimation method enables DyRadio to estimate the unknown measurement noise level and improve reconstruction performance compared with fixed noise-level input.

\subsubsection{Ablation Study}
To evaluate the contributions of the key components in DyRadio, we conduct ablation experiments by removing Stage I, temporal attention, and noise-conditioned observation fusion, respectively. As shown in Table~\ref{tab:ablation}, the full model consistently achieves the best performance across all evaluated settings. Removing Stage I leads to the most significant degradation, with PSNR reductions of $7.07$--$8.28$ dB, highlighting the importance of propagation-guided features for RM reconstruction. Removing temporal attention also reduces PSNR by $5.43$--$7.33$ dB, confirming the contribution of cross-frame information exchange to dynamic RM reconstruction. 
Furthermore, the performance gap between the
full model and the variant without noise-conditioned observation
fusion widens at higher measurement noise levels. At $p=2\%$ and $\sigma_{\mathrm{obs}}=0.09$, PSNR decreases by $3.01$ dB, while SSIM drops from $0.9785$ to $0.9429$. These results demonstrate the effectiveness of noise-conditioned observation fusion in improving reconstruction robustness under noisy observations.

\subsubsection{Inference Efficiency}

We evaluate the inference efficiency of different methods
on a single NVIDIA RTX PRO 6000 GPU.
DyRadio and RadioMapMotion generates $16$ frames per inference, whereas RadioUNet, RadioDiff, and PhyRMDM perform frame-wise inference with a batch size of $16$.
As shown in Table~\ref{tab:inference_efficiency},
DyRadio achieves superior reconstruction performance
while maintaining an inference time comparable to those
of diffusion-based RadioDiff and PhyRMDM methods.
RadioDiff has the largest parameter count among the
compared methods, partly because its reported count
includes the VAE module.

Noise estimation takes $1.842$ s ($11.9\%$ of the total inference time), introducing limited overhead while enabling adaptation to unknown measurement noise.

\begin{table}[!t]
    \centering
    \caption{Inference Efficiency of Different RM Reconstruction Methods.}
    \label{tab:inference_efficiency}
    \footnotesize
    \setlength{\tabcolsep}{3pt}
    \renewcommand{\arraystretch}{1.08}
    \begin{tabular*}{\columnwidth}{@{\extracolsep{\fill}}lccc@{}}
        \toprule
        Method / Stage &
        \shortstack{Parameters\\(M)} &
        \shortstack{Inference Time\\(s)} &
        \shortstack{Peak GPU Memory\\(GiB)} \\
        \midrule
        RadioUNet       & 13.28  & 0.072  & 0.370 \\
        RadioMapMotion  & 2.70   & 0.138  & 0.200 \\
        RadioDiff       & 279.10 & 11.625 & 2.140 \\
        PhyRMDM         & 66.47  & 18.445 & 2.460 \\
        \midrule
        DyRadio & & & \\
        \quad Noise Estimation
                        & -- & 1.842 & 2.949 \\
        \quad RM Reconstruction
                        & -- & 13.600 & 2.943 \\
        \quad \textbf{Complete Pipeline}
                        & \textbf{186.63}
                        & \textbf{15.442}
                        & \textbf{2.949} \\
        \bottomrule
    \end{tabular*}

    \vspace{1mm}
    \parbox{\columnwidth}{
        \footnotesize
        \textit{Note:} All inference times are measured for
        96-frame sequences. 
    }
\end{table}

\section{Conclusion}
\label{sec:conclusion}
In this work, we have proposed DyRadio, a framework integrating a noise-conditioned spatiotemporal DM with an iterative noise-level estimation method for dynamic RM reconstruction from sparse observations. In DyRadio, by integrating propagation-prior features with spatiotemporal attention, the DM jointly reconstructs consecutive RMs, while the noise estimation method uses the DM's learned RM prior to iteratively predict RMs in blind areas and update the noise estimate. Subsequently, noise-modulated cross-attention adaptively fuses sparse observations under varying noise levels, enabling robust reconstruction. Experimental results have demonstrated that DyRadio achieves superior reconstruction accuracy and robustness against measurement noise compared with existing methods. Moreover, the proposed noise estimation method accurately estimates unknown noise levels and enables effective noise-conditioned reconstruction. Future work will extend DyRadio to diverse dynamic environments beyond urban traffic and explore how reconstructed RMs can be applied in wireless-sensing-aware embodied tasks.


\appendix[Evaluation Metrics]

\label{app:evaluation_metrics}

We evaluate RM reconstruction using MSE, NMSE, PSNR, and
SSIM, and noise estimation using the average estimated noise,
MAE, and relative bias. The reconstruction metrics are computed
over each complete $T$-frame RM sequence and averaged equally
across test sequences. Let $X_t(h,w)$ and $\widehat X_t(h,w)$
denote the ground-truth and reconstructed RM values at location
$(h,w)$ in frame $t$, where $H$ and $W$ are the spatial
dimensions.

\textit{1) MSE:}
The average squared reconstruction error is
$\mathrm{MSE}=(THW)^{-1}\sum_{t,h,w}
(\widehat X_t(h,w)-X_t(h,w))^2$.

\textit{2) NMSE:}
The squared reconstruction error normalized by the
ground-truth RM energy is
$\mathrm{NMSE}=
\frac{\sum_{t,h,w}(\widehat X_t(h,w)-X_t(h,w))^2}
{\sum_{t,h,w}X_t(h,w)^2}$.

\textit{3) PSNR:}
PSNR evaluates reconstruction fidelity based on the ratio
of the maximum signal power to MSE:
$\mathrm{PSNR}=10\log_{10}(\mathrm{MAX}^2/\mathrm{MSE})$.

\textit{4) SSIM:}
SSIM evaluates local luminance, contrast, and structural
similarity between reconstructed and ground-truth RMs
~\cite{wang2004ssim}:
$\mathrm{SSIM}(\mathbf a,\mathbf b)
=
\frac{(2\mu_a\mu_b+C_1)(2\sigma_{ab}+C_2)}
{(\mu_a^2+\mu_b^2+C_1)(\sigma_a^2+\sigma_b^2+C_2)}$,
where $\mathbf a$ and $\mathbf b$ are corresponding local
patches, $\mu_a$ and $\mu_b$ are their means, $\sigma_a^2$
and $\sigma_b^2$ are their variances, $\sigma_{ab}$ is their
covariance, and $C_1$ and $C_2$ are stabilization constants.
Lower MSE and NMSE values, together with higher PSNR and
SSIM values, indicate better reconstruction quality.

\textit{5) Average estimated noise:}
Let $\hat{\sigma}_{\mathrm{obs}}^{(i)}$ denote the
noise estimate for the $i$-th test sequence. The average
estimated noise level is
$\bar{\hat{\sigma}}_{\mathrm{obs}}
=N^{-1}\sum_{i=1}^{N}\hat{\sigma}_{\mathrm{obs}}^{(i)}$.

\textit{6) MAE:}
The average absolute error in noise estimation is
$\mathrm{MAE}=N^{-1}\sum_{i=1}^{N}
|\hat{\sigma}_{\mathrm{obs}}^{(i)}-\sigma_{\mathrm{obs}}|$.

\textit{7) Relative Bias:}
Relative bias measures the percentage deviation of the mean
estimated noise level from the ground-truth value:
$\mathrm{Relative\ Bias}
=100\%\,(\bar{\hat{\sigma}}_{\mathrm{obs}}-\sigma_{\mathrm{obs}})
/\sigma_{\mathrm{obs}}$.

\bibliographystyle{IEEEtran}
\bibliography{main}
\end{document}